\documentclass{article}

 \usepackage[preprint]{neurips_2026}

\usepackage[utf8]{inputenc} 
\usepackage[T1]{fontenc}    
\usepackage{hyperref}       
\usepackage{url}            
\usepackage{booktabs}       
\usepackage{amsfonts}       
\usepackage{nicefrac}       
\usepackage{microtype}      
\usepackage{xcolor}         
\usepackage{amsmath}        
\usepackage{multirow}
\usepackage{makecell}
\usepackage{graphicx}
\usepackage{tabularx}
\usepackage{makecell}
\usepackage{pifont}
\usepackage{amsthm}
\usepackage{subcaption}
\usepackage{textcomp}
\usepackage[most]{tcolorbox}
\usepackage{fancyvrb}
\usepackage{caption}
\usepackage{wrapfig}   
\usepackage{fontawesome5}

\newcommand{\cmark}{\textcolor{green!60!black}{\ding{51}}}   
\newcommand{\xmark}{\textcolor{red}{\ding{55}}}     

\usepackage{enumitem}

\title{STT-Arena: A More Realistic Environment for Tool-Using with Spatio-Temporal Dynamics}

\author{
 \textbf{Tingfeng Hui\textsuperscript{1,3}},
 \textbf{Hao Xu\textsuperscript{4}},
  \textbf{Pengyu Zhu\textsuperscript{3}}, \\
 \textbf{Hongsheng Xin\textsuperscript{4}},
 \textbf{Kun Zhan\textsuperscript{4}},
 \textbf{Sen Su\textsuperscript{3}}, \textbf{Chunxiao Liu\textsuperscript{5}\thanks{The corresponding author}}, \textbf{Ning Miao\textsuperscript{1,2}}
\\
\\
 \textsuperscript{1}Hong Kong Institute of AI for Science, City University of Hong Kong \\
 \textsuperscript{2}Department of Data Science, City University of Hong Kong \\
 \textsuperscript{3}Beijing University of Posts and Telecommunications \\
 \textsuperscript{4}Li Auto Inc. \textsuperscript{5}Independent Researcher
}

\begin{document}

\maketitle

\begin{center}
    \href{https://github.com/Miaow-Lab/STT-Arena}{\faGithub \quad STT-Arena}\quad
    \href{https://huggingface.co/datasets/Miaow-Lab/STT-Arena}{%
        \includegraphics[height=1em]{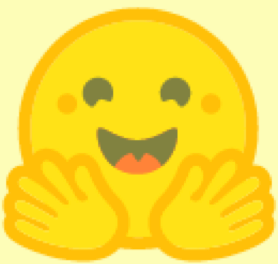} \quad Datasets
    }\quad
    \href{https://huggingface.co/Miaow-Lab/STT-Agent-SFT}{%
        \includegraphics[height=1em]{figures/huggingface.png} \quad STT-Agent-SFT
    }\quad
    \href{https://huggingface.co/Miaow-Lab/STT-Agent-RL}{%
        \includegraphics[height=1em]{figures/huggingface.png} \quad STT-Agent-RL
    }
\end{center}
\vspace{1em}

\begin{abstract}
Large language models (LLMs) deployed in real-world agentic applications must be capable of replanning and adapting when mid-task disruptions invalidate their prior decisions. Existing dynamic benchmarks primarily measure whether LLMs can detect temporal changes in a timely manner, leaving the complementary challenge of adaptive replanning under spatio-temporal dynamics largely unexplored. We introduce STT-Arena (Spatio-Temporal Tool-Use Arena), a benchmark of 227 high-quality interactive tasks spanning nine spatio-temporal conflict types and four solvability levels. Each task is grounded in a realistic, executable environment equipped with injected spatio-temporal triggers that can abruptly invalidate an ongoing plan, forcing the model to detect the state shift and construct a revised execution strategy. Extensive evaluation of frontier LLMs reveals that even the SOTA proprietary models, including Claude-4.6-Opus, achieves less than 40\% overall accuracies, highlighting the fundamental difficulty of spatio-temporal dynamic reasoning. 
Systematic analysis of failure trajectories uncovers three recurring error modes of existing models: Stale-State Execution, Misdiagnosis of Dynamic Triggers, and Missing Post-Adaptation Verification. 
Guided by these findings, we propose an iterative trajectory refinement technique that eliminates these failure patterns from training data, and combine it with online RL to produce STT-Agent-4B which outperforms frontier LLMs on STT-Arena.

\end{abstract}


\section{Introduction}

Large language models (LLMs) based agents are increasingly deployed in real-world commercial applications, including airline reservation systems, clinical consultation services, etc \citep{DBLP:journals/corr/abs-2505-09388, DBLP:journals/corr/abs-2602-02276, DBLP:journals/corr/abs-2602-15763, DBLP:journals/corr/abs-2510-18855, DBLP:journals/corr/abs-2601-05808, DBLP:journals/corr/abs-2602-09372}. In these settings, LLMs must interact with external environments to retrieve information and execute multi-step operations. While early studies \citep{DBLP:conf/emnlp/LiZ000YLHL23, DBLP:conf/iclr/QinLYZYLLCTQZHT24, DBLP:journals/corr/abs-2406-12045, DBLP:journals/corr/abs-2509-26490} focus on static environment benchmarks with fixed interfaces and predictable outputs, more recent works such as GAIA-2 \citep{DBLP:journals/corr/abs-2602-11964}, Real-Time Reasoning Gym \citep{DBLP:journals/corr/abs-2511-04898}, and Timely Machine \citep{DBLP:journals/corr/abs-2601-16486} introduce continuously evolving environments. These benchmarks emphasize real-time environmental change and measure how rapidly an LLM can respond to external dynamics, treating responsiveness as the primary indicator of competence under non-stationary conditions.

\begin{wrapfigure}{r}{0.5\textwidth}
    \centering
    \vspace{-1.0em}
    \includegraphics[width=\linewidth]{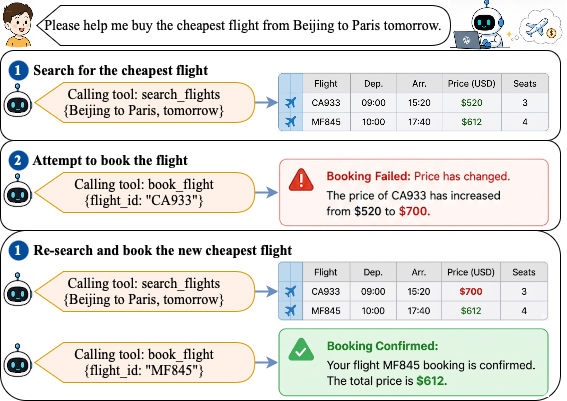}
    \caption{Adaptive replanning in spatio-temporal environments: A mid-task price change invalidates the plan, prompting detection of updated prices and reselection of the optimal flight.}
    \label{fig:st-intro}
\end{wrapfigure}

Despite these advances, existing dynamic benchmarks focus on timely completion under continuously evolving environments. In this work, we address a complementary dimension: \textbf{\textit{adaptive replanning and recovery}}, the ability to abandon a failed plan and reconstruct an alternative multi-step strategy when a sudden spatio-temporal change invalidates prior decisions. As illustrated in Figure \ref{fig:st-intro}, consider an LLM tasked with purchasing the cheapest available flight ticket. Midway through execution, ticket prices shift due to spatio-temporal dynamics. Beyond simply detecting the change, a capable LLM must reassess its prior decisions and construct a revised plan to complete the task correctly. We argue that the ability to replan and reconsider under mid-task environmental shifts is essential for trustworthy real-world deployment.

To systematically characterize the environmental dynamics that necessitate such replanning, we identify three fundamental axes along which real-world conditions evolve. \textbf{\textit{Temporal evolution}} refers to state changes that unfold over time: for instance, seat availability changes continuously as passengers complete bookings during online check-in. \textbf{\textit{Spatial dependency}} captures how environmental conditions vary with geographic context: delivery services, for example, are bound by predefined service zones, and a change in location may render certain operations inaccessible or alter operational constraints. \textbf{\textit{Spatio-temporal dynamics}} arise when a task is jointly governed by both dimensions: during rush hour, traffic congestion propagates across different locations at different times, making it unreliable to plan a route without accounting for their coupled interaction.

\begin{figure}
    \centering
    \includegraphics[width=0.98\linewidth]{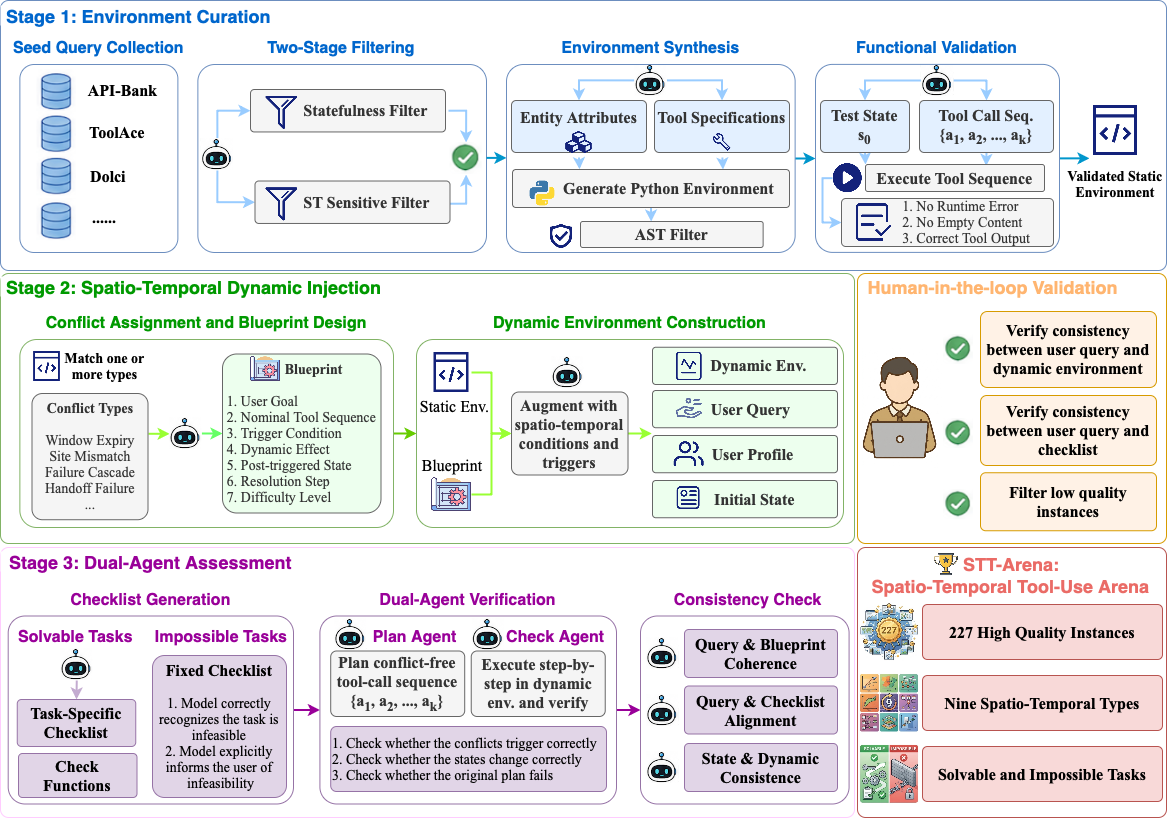}
    \caption{Overview of the STT-Arena construction pipeline. The pipeline consists of three stages: (1) Environment Curation, (2) Spatio-Temporal Dynamic Injection, and (3) Dual-Agent Assessment. Then we apply human-in-the-loop review to produce the final 227 benchmark instances.}
    \label{fig:pipeline}
    \vspace{-1.5em}
\end{figure}

Grounded in these three axes, we introduce \textbf{STT-Arena} (Spatio-Temporal Tool-Use Arena), a dynamic and interactive benchmark designed to evaluate the replanning and adaptive tool-use capabilities of LLMs under spatio-temporal conditions (Table \ref{tab:feature_matrix} shows the comparison of existing work). Built upon a diverse collection of authentic scenarios, STT-Arena makes three core contributions:

\begin{itemize}[leftmargin=1.5em, itemsep=0.4em, topsep=0.2em]

\item \textbf{A fine-grained, spatio-temporally centered task taxonomy.} We curate a comprehensive set of authentic tool-use tasks organized into 3 major categories and 9 subcategories that explicitly capture temporal evolution, spatial dependency, and their coupled interactions. Each task is paired with a dedicated executable environment that faithfully and controllably simulates real-world spatio-temporal dynamics across diverse domains. Tasks are further stratified by solvability: \textit{solvable} tasks are divided into three difficulty levels (\textit{Easy}, \textit{Medium}, and \textit{Hard}), where the LLM must identify a feasible action sequence under evolving conditions; \textit{impossible} tasks require the LLM to recognize that no valid completion path exists and to correctly report task infeasibility. Detailed information about the nine subcategories and four levels can be found in Tables \ref{tab:conflict_taxonomy_compact} and \ref{tab:difficulty_levels}.

\item \textbf{A scalable, interactive infrastructure for spatio-temporal environment simulation.} As shown in Figure \ref{fig:pipeline}, we introduce a dynamic simulation framework built around three core components: environment curation, spatio-temporal dynamic injection, and dual-agent assessment. This design enables the systematic evaluation and targeted improvement of model tool-use capabilities under dynamically changing conditions, offering a reproducible, extensible, and high-fidelity testbed that closely mirrors real-world deployment scenarios.

\item \textbf{Extensive benchmarking results and an efficient training paradigm for spatio-temporal tool augmentation.} As shown in Figure \ref{fig:main_results}, we evaluate closed-source and open-source LLMs at scale, exposing critical deficiencies in their ability to replan and operate under dynamic tool-use constraints. Guided by analyses of recurring failure modes, we introduce an iterative trajectory refinement technique that post-processes training trajectories by reordering, deleting, or modifying tool-call blocks to eliminate inefficient interaction patterns. Building on 2,212 refined trajectories for SFT and a set of verifiable dynamic tasks for online RL, we release \textbf{STT-Agent-4B}, which achieves 27.17\% on STT-Arena, matching the performance of GLM-5.1.

\end{itemize}

\begin{table}[htbp]
\centering
\caption{Feature comparison of STT-Arena against existing agentic tool-use benchmarks across three dimensions: evaluation protocol, environment type, and Training Data.}
\label{tab:feature_matrix}
\resizebox{\textwidth}{!}{
\begin{tabular}{l ccc cccccc cc}
\toprule
\multirow{3}{*}{\textbf{Benchmark}} & \multicolumn{3}{c}{\textbf{Evaluation Protocol}} & \multicolumn{6}{c}{\textbf{Environment Type}} & \multicolumn{2}{c}{\textbf{Training Data}} \\
\cmidrule(lr){2-4} \cmidrule(lr){5-10} \cmidrule(lr){11-12}
& \textbf{Tool} & \textbf{State} & \textbf{LLM} & \textbf{No} & \textbf{Static} & \textbf{Realistic} & \multirow{2}{*}{\textbf{Temporal}} & \multirow{2}{*}{\textbf{Spatial}} & \textbf{Spatio} & \multirow{2}{*}{\textbf{SFT}} & \multirow{2}{*}{\textbf{RL}} \\
& \textbf{Matching} & \textbf{Alignment} & \textbf{Judgement} & \textbf{Env.} & \textbf{Env.} & \textbf{Time} & & & \textbf{Temporal} & & \\
\midrule
API-Bank         & \cmark & \cmark & \xmark & \cmark & \xmark & \xmark & \xmark & \xmark & \xmark & \cmark & \xmark \\
ToolBench        & \xmark & \cmark & \xmark & \cmark & \xmark & \xmark & \xmark & \xmark & \xmark & \cmark & \xmark \\
StableToolBench  & \xmark & \cmark & \xmark & \cmark & \xmark & \xmark & \xmark & \xmark & \xmark & \xmark & \xmark \\
MCP-Bench        & \cmark & \xmark & \cmark & \cmark & \xmark & \xmark & \xmark & \xmark & \xmark & \xmark & \xmark \\
MCPToolBench++  & \cmark & \cmark & \cmark & \cmark & \xmark & \xmark & \xmark & \xmark & \xmark & \xmark & \xmark \\
ToolTalk         & \cmark & \cmark & \xmark & \cmark & \xmark & \xmark & \xmark & \xmark & \xmark & \xmark & \xmark \\
\midrule
BFCL-v4          & \cmark & \cmark & \xmark & \xmark & \cmark & \xmark & \xmark & \xmark & \xmark & \xmark & \xmark \\
$\tau$-Bench     & \xmark & \cmark & \xmark & \xmark & \cmark & \xmark & \xmark & \xmark & \xmark & \xmark & \xmark \\
$\tau^2$-Bench   & \xmark & \cmark & \xmark & \xmark & \cmark & \xmark & \xmark & \xmark & \xmark & \xmark & \xmark\\
ToolSandBox      & \cmark & \cmark & \xmark & \xmark & \cmark & \xmark & \xmark & \xmark & \xmark & \xmark & \xmark \\
ACEBench         & \cmark & \cmark & \xmark & \xmark & \cmark & \xmark & \xmark & \xmark & \xmark & \xmark & \xmark \\
ToolAthlon          & \xmark & \cmark & \xmark & \xmark & \cmark & \xmark & \xmark & \xmark & \xmark & \xmark & \xmark \\
VitaBench & \xmark & \xmark & \cmark & \xmark & \cmark & \xmark & \xmark & \xmark & \xmark & \xmark & \xmark \\
\midrule
TCP              & \xmark & \cmark & \xmark & \xmark & \xmark & \cmark & \xmark & \cmark & \xmark & \xmark & \xmark \\
Timely-Eval      & \xmark & \cmark & \cmark & \xmark & \xmark & \cmark & \xmark & \xmark & \xmark & \cmark & \cmark \\
GAIA-2           & \cmark & \xmark & \cmark & \cmark & \xmark & \cmark & \xmark & \xmark & \xmark & \xmark & \xmark \\
RTR Gym          & \xmark & \cmark & \xmark & \cmark & \xmark & \cmark & \xmark & \cmark & \xmark & \xmark & \xmark \\
\midrule
\textbf{STT-Arena (Ours)} & \xmark & \cmark & \cmark & \xmark & \xmark & \xmark & \cmark & \cmark & \cmark & \cmark & \cmark \\
\bottomrule
\end{tabular}
}
\vspace{-1.0em}
\end{table}

\section{Spatio-Temporal Tool-Use Arena}
In this section, we present STT-Arena, a benchmark for evaluating LLMs in tool-use environments that evolve autonomously over time and space. Each task includes spatio-temporal triggers that modify the environment state or tool availability when certain conditions are met, forcing the model to detect changes and replan accordingly. Each task is created using a three-stage approach.

\subsection{Task Formulation}
STT-Arena formalizes each instance as a tuple $\mathcal{T} = (\mathcal{E}, \Phi, u, q, \mathrm{CL})$, where $\mathcal{E}$ denotes the environment, which encompasses a set of states $\mathcal{S}$ and available tools $\mathcal{A}$; $\Phi = {(\phi, c_\phi, e_\phi)}$ is spatio-temporal triggers, each with a condition $c_\phi: \mathcal{S} \times \mathcal{X} \to \{0,1\}$ depending on the state and the spatio-temporal context $\mathcal{X}$ and an effect $e_\phi: \mathcal{S} \to \mathcal{S}$ that modifies the state or tool availability when $c_\phi$ holds; $u$ is the user profile encoding preferences and constraints; $q$ is the user query; and $\mathrm{CL}$ is a checklist of necessary conditions for task success. The LLM receives $q$ and issues a sequence of tool calls from $\mathcal{A}$. The model may also query a passive user simulator to clarify ambiguities or confirm state changes, but no additional information is provided beyond $q$ and $u$. At any step, if the current state and context satisfy $c_\phi$ for some trigger $\phi \in \Phi$, the corresponding effect $e_\phi$ is applied autonomously. Such changes can invalidate the model's previous plan, forcing it to re-plan and adapt. After the model terminates, the final state is evaluated against $\mathrm{CL}$ to obtain the pass@1 rate of the task.

\subsection{STT-Arena Construction Pipeline}
To construct STT-Arena, we design a three-stage pipeline that systematically transforms real-world user requests into executable and rigorously validated spatio-temporal dynamic tasks. The pipeline begins by curating reliable static environments including entity states and tools. Then, it injects controlled spatio-temporal conflicts to create dynamic task instances with solvable and impossible categories across nine spatio-temporal types. Finally, we employ dual-agent assessment and human review to ensure that each instance is realistic, internally consistent, and evaluation-ready. The specific cases of each stage and step can be found in Appendix \ref{sec:case_pipeline}.

\subsubsection{Stage 1: Environment Curation.}
The first stage, inspired by \citep{DBLP:journals/corr/abs-2601-05808}, constructs a library $\mathcal{E}_{\mathrm{static}}$ of validated, executable static environments that serve as the foundation for dynamic task generation.

\textbf{Seed Query Collection and Filtering.}
We collect real-world user queries from three sources: API-Bank \citep{DBLP:conf/emnlp/LiZ000YLHL23}, ToolAce \citep{DBLP:conf/iclr/Liu0ZHYL0GLY0WN25}, and Dolci \citep{olmo2025olmo3} to ensure that our benchmark is diverse and representative of real-world distributions. Each query passes through a two-stage filter using an LLM. The \textbf{statefulness filter} checks whether the query requires a persistent environment state across steps. The \textbf{spatio-temporal sensitivity filter} checks whether the outcome changes as a function of time or location. Only queries that pass both filters are retained. The detailed prompts can be found in Appendix \ref{sec:two_stage_filter_prompt}.

\textbf{Environment Synthesis.}
For each qualifying query, we first prompt an LLM to infer the latent environmental information, including an environment summary and a detailed introduction. Based on this inferred information, we generate the corresponding entity attributes and tool specifications. The entity states and tool logic are then implemented separately as executable Python classes. Finally, we concatenate the entity attribute classes and the tool environment classes, and pass the combined code through an AST filter to eliminate unsafe or non-deterministic constructs, yielding a candidate environment $e \in \mathcal{E}_{\mathrm{candidate}}$. The detailed prompts can be found in Appendix \ref{sec:environment_synthesis_prompt}.

\textbf{Functional Validation.}
Each candidate environment $e$ is validated by a tool-calling LLM. The LLM generates diverse test configurations, each consisting of an initial environment state $s_0$ and a sequence of tool calls $\{a_1, \dots, a_k\}$. The environment executes this sequence; if any execution produces a runtime error, returns empty content, or otherwise fails to produce the expected tool outputs, the environment is discarded. Only environments that pass all test configurations without error are promoted to $\mathcal{E}_{\mathrm{static}}$. The detailed prompts can be found in Appendix \ref{sec:functional_validation_prompt}.

\subsubsection{Stage 2: Spatio-Temporal Dynamic Injection}
The second stage mainly construct spatio-temporal dynamic environments $\mathcal{E}_{\mathrm{dynamic}}$ and tasks.

\textbf{Conflict Assignment and Blueprint Design.}
For each $e \in \mathcal{E}_{\mathrm{static}}$, we select one or more semantically compatible conflict types from a predefined set $\mathcal{C}$ of nine spatio-temporal dynamic types (Table \ref{tab:conflict_taxonomy_compact}). We then generate a blueprint $B$ with difficulty level $d$ (easy, medium, hard, and impossible) to generate conflict stories. Here, the blueprint acts as a generative contract that enforces internal consistency across all downstream modules, which includes: (1) User goal and user profile; (2) Nominal tool sequence $\{a_1, \dots, a_m\}$ that succeeds in the absence of conflict; (3) Conflict trigger condition $c_\phi$, which depends on the state and the spatio-temporal context $\mathcal{X}$; (4) Effect of conflict $e_\phi$, which modifies the state or tool availability when condition $c_\phi$ satisfied; (5) Expected post-trigger state, characterizing the revised environmental states or tool availability that the model needs to recognize; (6) Required resolution steps that the model must execute to resolve the conflict and regain the objective. The detailed prompt templates can be found in Appendix~\ref{sec:conflict_and_blueprint}.

\textbf{Dynamic Environment Construction.}
Given the blueprint $B$ and a static environment $e \in \mathcal{E}_{\mathrm{static}}$, our construction process involves three key steps:
First, we prompt an LLM to augment $e$ with the spatio-temporal conditions $\Phi$ specified in $B$, producing a dynamic environment $e_{\mathrm{dynamic}}$. Second, we synthesize the user query $q$ and user profile $u$ based on the goals and constraints defined in $B$; Finally, we establish a realistic initial state $s_0 \in \mathcal{S}_e$ following the specified task plan in $B$. The detailed prompt templates can be found in Appendix \ref{sec:dynamic_construction}.

\subsubsection{Stage 3: Dual-Agent Assessment}

The third stage focuses on validating the dynamic environments and tasks through dual-agent protocol.

\textbf{Checklist Generation.}
For each dynamic instance, we generate an evaluation checklist $\mathrm{CL}$ based on the blueprint $B$ and its difficulty level $d$. For feasible tasks (easy, medium, and hard), we prompt an LLM to produce a task-specific checklist $\mathrm{CL} = \{\mathrm{criterion}_1, \dots, \mathrm{criterion}_p\}$ enumerating the necessary and sufficient conditions for task success, along with a set of rule-based check functions $\mathcal{F} = \{f_1, \dots, f_p\}$, where each $f_j: \mathcal{S} \to \{0,1\}$ evaluates the final environment state against each criterion. For the impossible category, the checklist is fixed and consists of two criteria: (i) the model correctly recognizes that the dynamic task cannot be completed given the current environment state, and (ii) the model explicitly informs the user simulator of this infeasibility. The detailed prompt templates can be found in Appendix \ref{sec:checklist_generation}.

\textbf{Dual-Agent Verification.}
To ensure data validity, each instance undergoes a dual-agent verification protocol. First, a planning agent formulates an original tool call sequence $\left\{a_1^, \dots, a_m^,\right\}$ that assumes a conflict-free execution. Next, a checking agent executes this sequence in the dynamic environment $e_{\mathrm{dynamic}}$ , strictly verifying the process against three behavioral invariants: (i) the condition $c_\phi$ triggers exactly as scheduled in $B$; (ii) $e_\phi$ modifies the context or tool availability as specified; and (iii) the conflict successfully disrupts the original plan $\{a_1^, \dots, a_m^,\}$, preventing goal achievement, e.g., a tool returns an error or an expected state change does not occur. Finally, spatio-temporal dynamic environments and tasks that pass this verification are curated for use as benchmark and training data. Please refer to Appendix \ref{sec:dual_agent_validation} for detailed prompt templates.

\textbf{Consistency Check.}
Subsequent to dual-agent verification, we further perform a consistency check using LLMs and human annotations. Firstly, LLM-based auditor checks execution trajectory to verify that all artifacts are mutually coherent: the user query matches the blueprint, concrete mutations realize the claimed conflict semantics, the evaluation checklist covers key success and failure conditions, and the difficulty level aligns with the observed complexity. Prompt templates for this process can be found in Appendix \ref{sec:consistency_check}. Second, we manually filter candidate instances by validating the user query  $q$, the checklist $\mathrm{CL}$, and the alignment between the task description and environment dynamics. This human-in-the-loop validation results in the final STT-Arena, which consists of 227 high-quality instances. Detailed information and statistics are provided in Appendix \ref{sec:detailed_information_of_stt_arena}.

\subsection{Task Evaluation}
Each task is evaluated through interaction between the LLM and a passive user simulator. After the model terminates, the final state $s_T$ is recorded.
For feasible tasks (easy, medium, hard), each instance has a set of check functions $\mathcal{F} = \{f_1, \dots, f_p\}$ where $f_j: \mathcal{S} \to \{0,1\}$. The per-function outcome provides a fine-grained reward $R_{\text{fea}} = \frac{1}{p}\sum_{j=1}^p f_j(s_T)$.
For impossible tasks, we use an LLM-as-a-judge with a fixed two-item checklist: (i) the model recognizes infeasibility, (ii) it communicates this to the user. The judge produces binary verdicts $v_1, v_2 \in {0,1}$ on the whole trajectory, yielding a binary reward $R_{\text{imp}} = \mathbf{1}[v_1 = 1 \land v_2 = 1]$.

\begin{figure}
    \centering
    \includegraphics[width=0.98\linewidth]{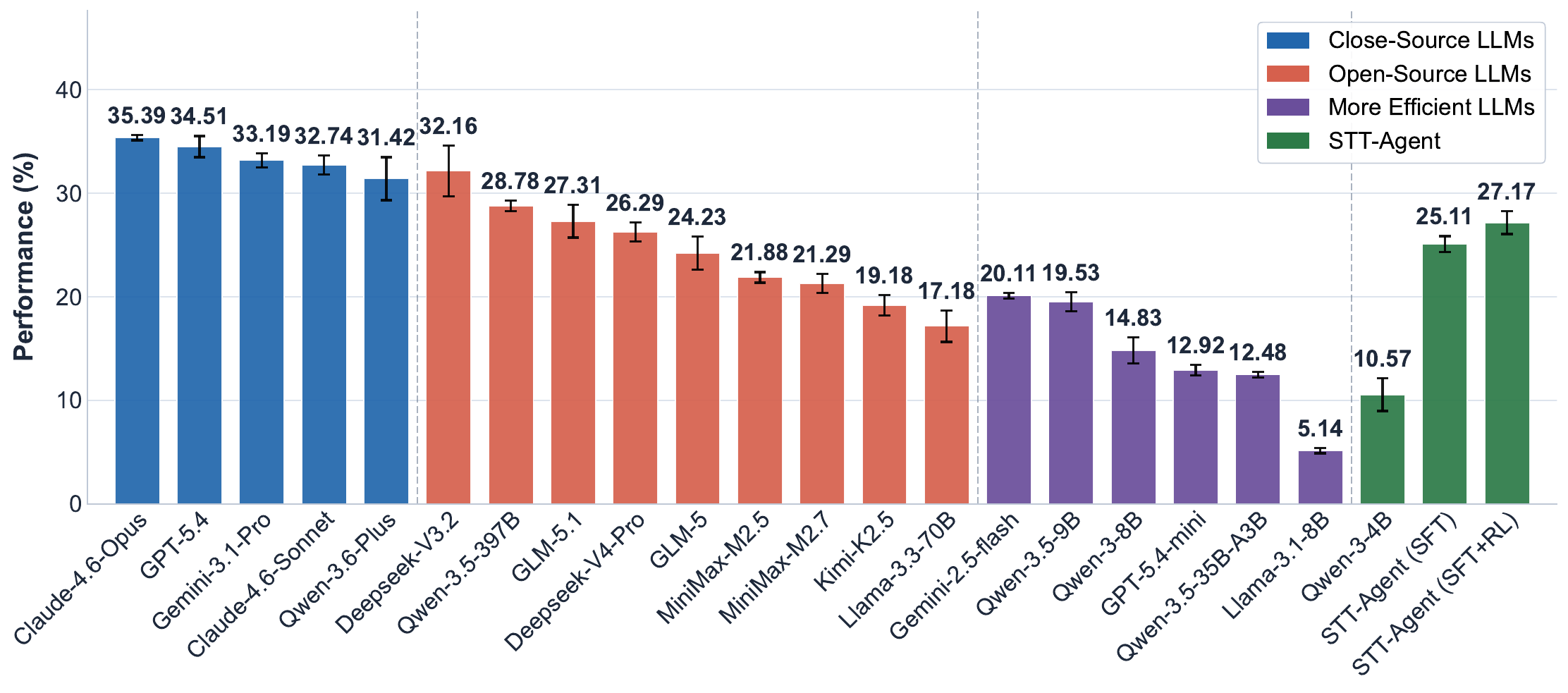}
    \caption{Overall Pass@1 performance of all evaluated models on STT-Arena. Results are grouped into four categories: closed-source LLMs, open-source LLMs, more efficient LLMs, and STT-Agent variants. Even the best-performing model, Claude-4.6-Opus, achieves only 35.39\%, underscoring the fundamental difficulty of spatio-temporal dynamic reasoning. STT-Agent-4B, despite having only 4B parameters, outperforms many open-source frontier models. Detailed results can be found in Table \ref{tab:main_results}.}
    \label{fig:main_results}
    \vspace{-1em}
\end{figure}

\section{Experiments}
\label{sec:experiments}

\subsection{Experimental Setup}
\textbf{Compared LLMs.} We benchmark the following LLMs on STT-Arena: (i) \textit{Closed-Source LLMs}, including GPT-5.4, Gemini-3.1-Pro, CLaude-4.6-Opus, Claude-4.6-Sonnet, and Qwen-3.6-Plus; (ii) \textit{Open-Source LLMs}, including GLM-5.1, GLM-5, Kimi-K2.5, MiniMax-M2.7, MiniMax-M2.5, Llama-3.3-70B, Qwen-3.5-397B-A17B, Deepseek-V3.2 and Deepseek-V4-Pro; (iii) \textit{More Efficient LLMs}, including GPT-5.4-mini, Gemini-2.5-flash, Llama-3.1-8B, Qwen-3.5-9B, Qwen-3.5-35B-A3B, and Qwen-3-8B; (iv) \textit{STT-Agent}, Qwen-3-4B (baseline) and STT-Agent with SFT and RL.

\textbf{Evaluation Metrics.} 
We adopt \textbf{Pass@1} as the primary evaluation metric. For the overall performance across solvable and impossible tasks, we compute a weighted average of the Pass@1 scores for each category, where the weight is proportional to the number of instances. Formally, 
\begin{equation}
    \text{Overall} = \alpha P_{\text{e}} + \beta P_{\text{m}} + \gamma P_{\text{h}} + \delta P_{\text{i}},
\end{equation}
where $P_{\text{e}}$, $P_{\text{m}}$, $P_{\text{h}}$, $P_{\text{i}}$ denote the Pass@1 scores for each category (easy, medium, hard, impossible, respectively), and the weighting coefficients $\alpha$, $\beta$, $\gamma$, $\delta$ are determined by sampling according to the corresponding number of instances in each level, with detailed values provided in Appendix \ref{sec:detailed_information_of_stt_arena}.

\textbf{Evaluation Details.} We set the maximum number of interaction turns to 50, perform three runs with a temperature of 0.7, and report the mean along with the standard deviation. We use Qwen-3.5-397B as the user simulator and the judgment model, with the temperature set to 0. Detailed system prompts used during evaluation can be found in the Appendix \ref{sec:stt_arena_evaluation}.

\subsection{Main Evaluation: STT-Arena Reveals Fundamental Gaps in Dynamic Reasoning}
\label{sec:main_results}

Figure~\ref{fig:main_results} and Table~\ref{tab:main_results} present the Pass@1 results across solvable and impossible tasks. Overall, current LLMs exhibit limited capabilities on STT-Arena, which can be summarized in three key observations:

\textbf{Overall performance is limited, highlighting fundamental task difficulty.} All evaluated models achieve limited performance on STT-Arena, with the best-performing model, Claude-4.6-Opus, reaching only 35.39\% overall, highlighting the fundamental difficulty of spatio-temporal dynamic reasoning. Performance consistently degrades as task difficulty increases from Easy to Hard, with all models exhibiting notable drops at the Hard level, confirming that long-horizon replanning under intertwined spatio-temporal constraints remains an open challenge for current LLMs.

\textbf{Closed-source models lead, while open-source models trail despite competitiveness.} Among closed-source LLMs, the Claude and GPT series lead the rankings. Open-source models show competitive but consistently lower performance, with Deepseek-V3.2 (32.16\%) being the strongest open-source contender yet still trailing the closed-source leaders by a non-trivial margin. This gap suggests that frontier closed-source models retain meaningful advantages in instruction following and adaptive decision-making under dynamic conditions.

\textbf{Efficient LLMs perform substantially worse, underscoring the critical role of model scale.} Efficient LLMs perform substantially worse than their frontier-scale counterparts. Models such as Llama-3.1-8B (5.14\%), Qwen-3.5-35B-A3B (12.48\%), and GPT-5.4-mini (12.92\%) lag far behind, indicating that model scale plays a critical role in handling the complex replanning demands of STT-Arena, and that parameter-efficient architectures alone are insufficient to address spatio-temporal dynamic reasoning without targeted training.

\begin{figure}[t]
    \centering
    \begin{minipage}{0.47\textwidth}
        \centering
        \includegraphics[width=\textwidth]{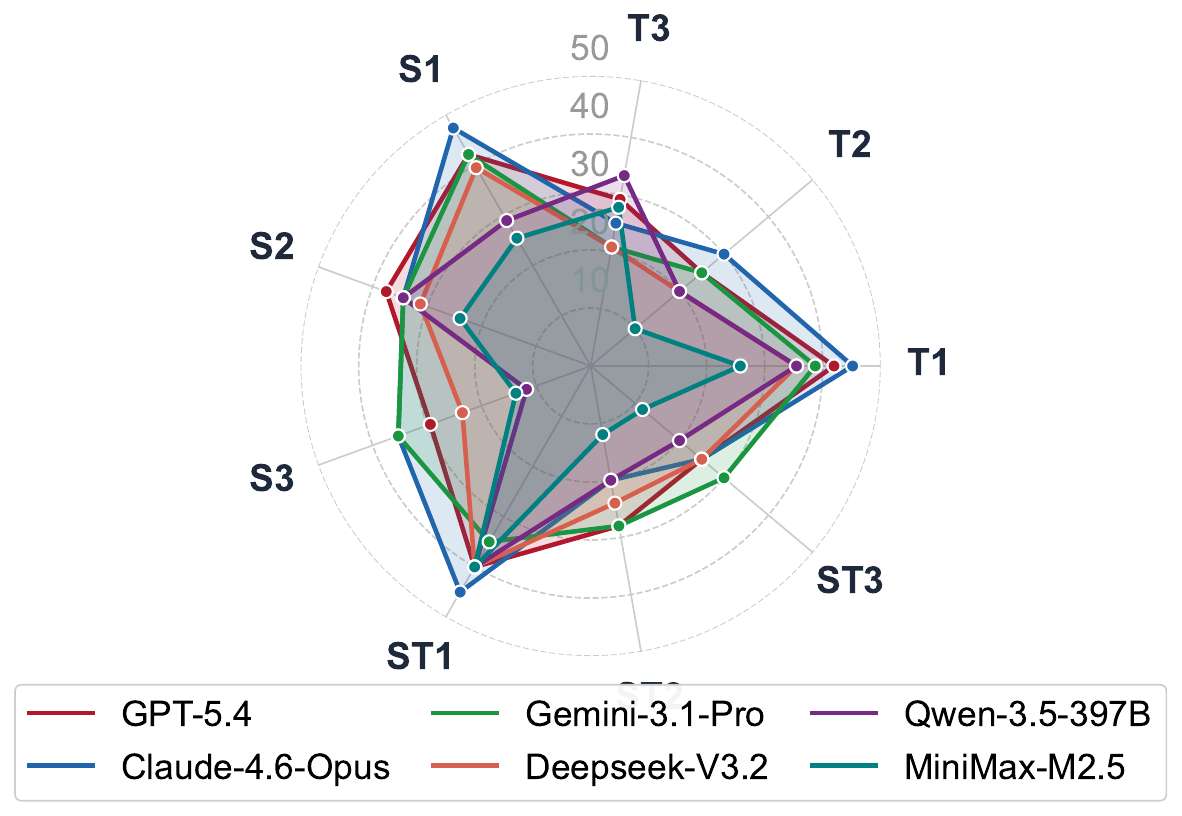}
        \captionof{figure}{Pass@1 performance across the nine spatio-temporal conflict subtypes.}
        \label{fig:subtypes}
    \end{minipage}
    \hfill
    \begin{minipage}{0.47\textwidth}
        \centering
        \includegraphics[width=\textwidth]{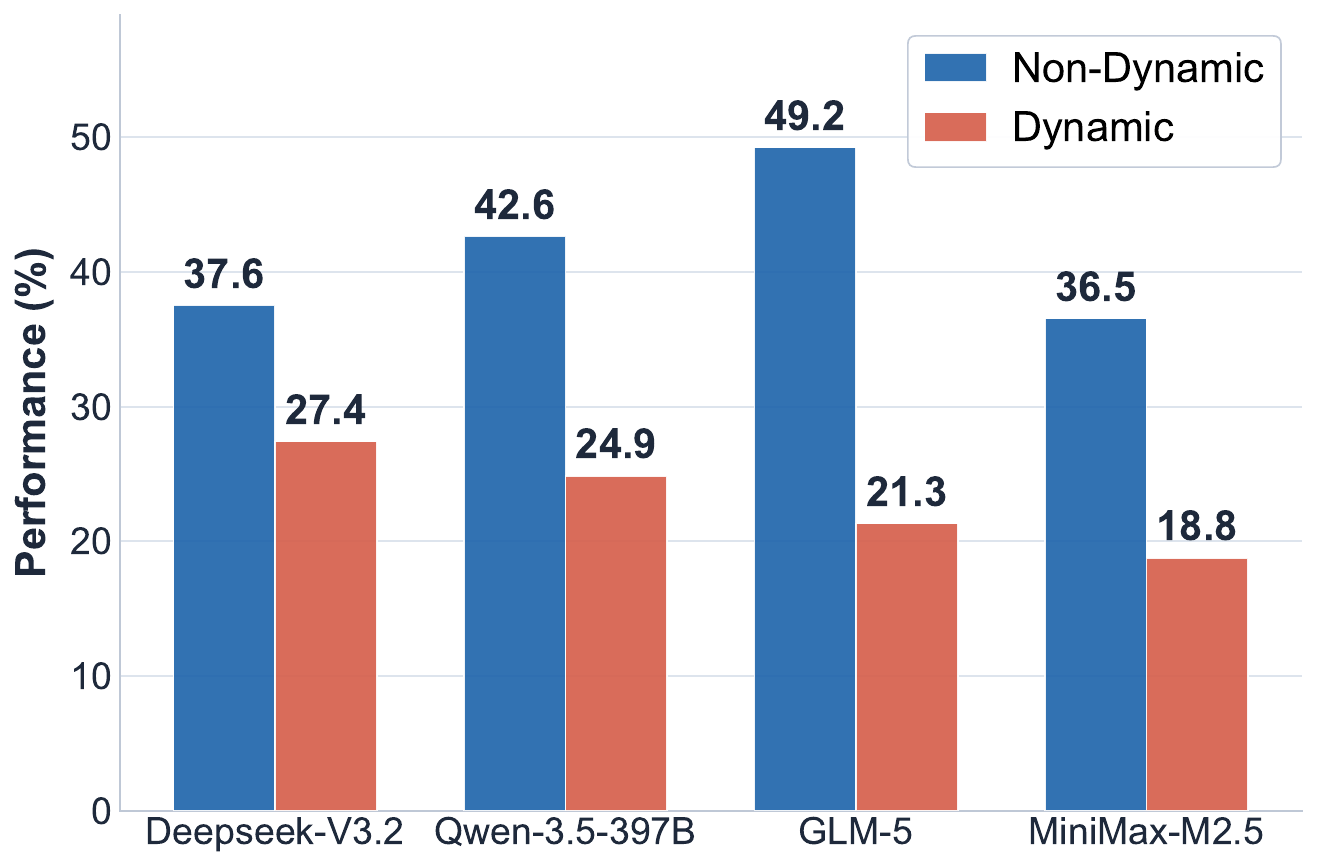}
        \captionof{figure}{Performance gap between dynamic (STT-Arena) and non-dynamic environments.}
        \label{fig:dynamic}
    \end{minipage}
    \vspace{-1.5em}
\end{figure}

\subsection{Key Analyses: What Makes STT-Arena Hard?}

To understand the origins of these performance gaps, we conduct fine-grained analyses along two dimensions, yielding the following insights:

\textbf{Task structure governs failure modes.} Figure~\ref{fig:subtypes} shows performance per conflict type defined in Table~\ref{tab:conflict_taxonomy_compact}. Across all models, T1 (window expiry) and S1 (site mismatch) are consistently high, while T2 (priority reorder) and S3 (route restriction) are consistently low. This indicates that within pure temporal or spatial dynamics, tasks involving simple deadline tracking or location mismatch are manageable, but those requiring reordering of priorities or enforcement of route constraints pose universal difficulty. Strikingly, ST1 (resource shift) achieves stable and high performance across every model. In contrast, ST2 (failure cascade) and ST3 (handoff failure) drop sharply. This contrast reveals that spatio-temporal coupling is not inherently difficult: simple resource reallocation (ST1) is well handled, but once the coupling involves cascading dependencies (ST2) or misaligned handoffs across time and space (ST3), all current LLMs break down, exposing a fundamental blind spot in multi-step causal reasoning under intertwined dynamics.

\textbf{Dynamics expose brittleness concealed in static evaluations.} As shown in Figure~\ref{fig:dynamic}, we compare model performance under a static benchmark constructed by removing spatio-temporal triggers and reconstructing the original checklist against STT-Arena. Across all four models, introducing dynamics consistently reduces performance, confirming that spatio-temporal evolution imposes a universal difficulty that current LLMs are not yet equipped to handle. More importantly, the relative ordering of models changes substantially between the two conditions, indicating that static evaluation alone is not sufficient for assessing robustness under realistic environmental shifts. These results suggest that high performance on conventional tool-use benchmarks may come at the cost of overfitting to fixed patterns, and that spatio-temporal stress testing is essential for assessing true deployment readiness.

\subsection{Further Ablations: Probing Model Capabilities and Design Choices}

We further investigate whether current limitations can be mitigated through algorithmic or architectural interventions, leading to three key findings:

\begin{figure}[htbp]
    \centering
    \begin{minipage}{0.47\textwidth}
        \centering
        \includegraphics[width=\textwidth]{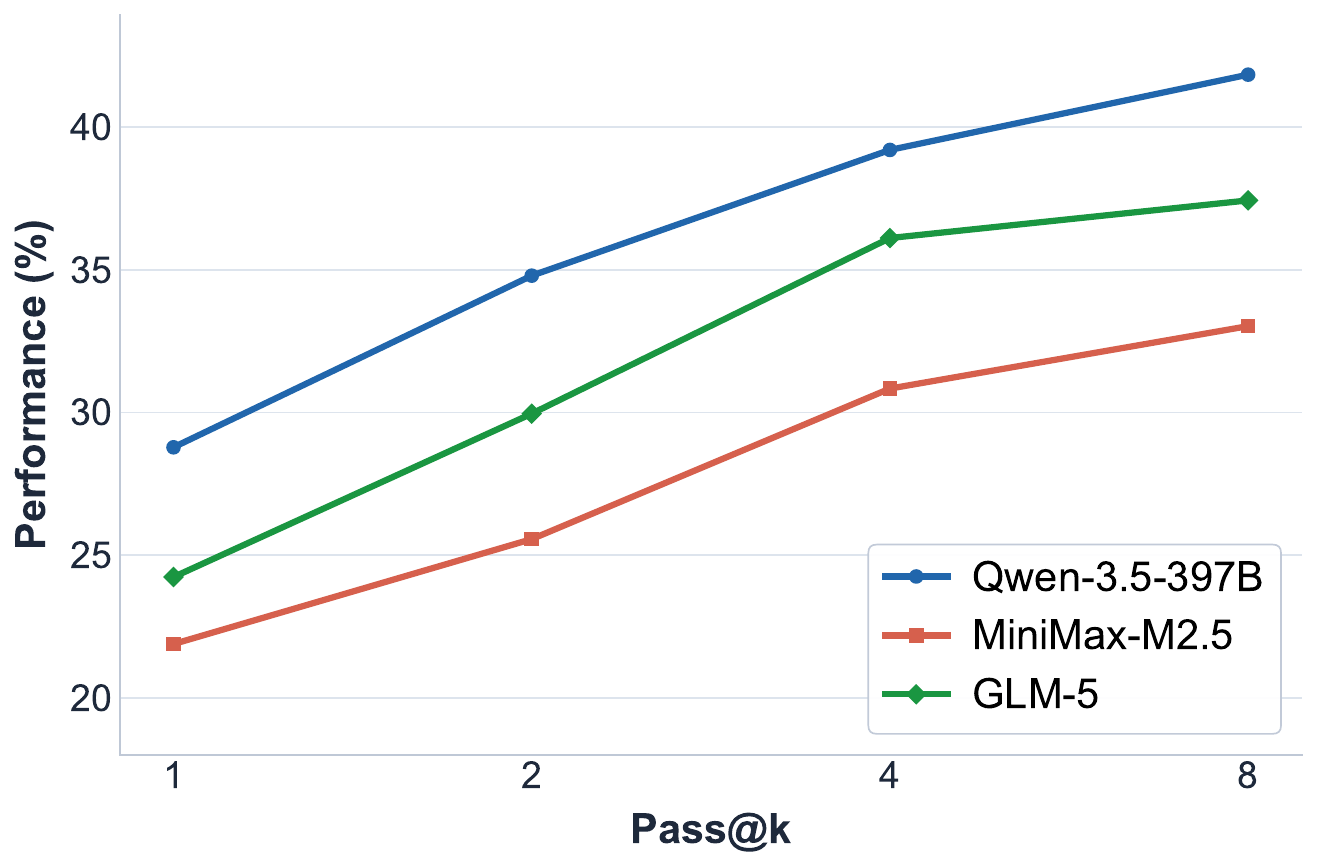}
        \captionof{figure}{Test-time scaling via Pass@k rate.}
        \label{fig:testtime}
    \end{minipage}
    \hfill
    \begin{minipage}{0.47\textwidth}
        \centering
        \includegraphics[width=\textwidth]{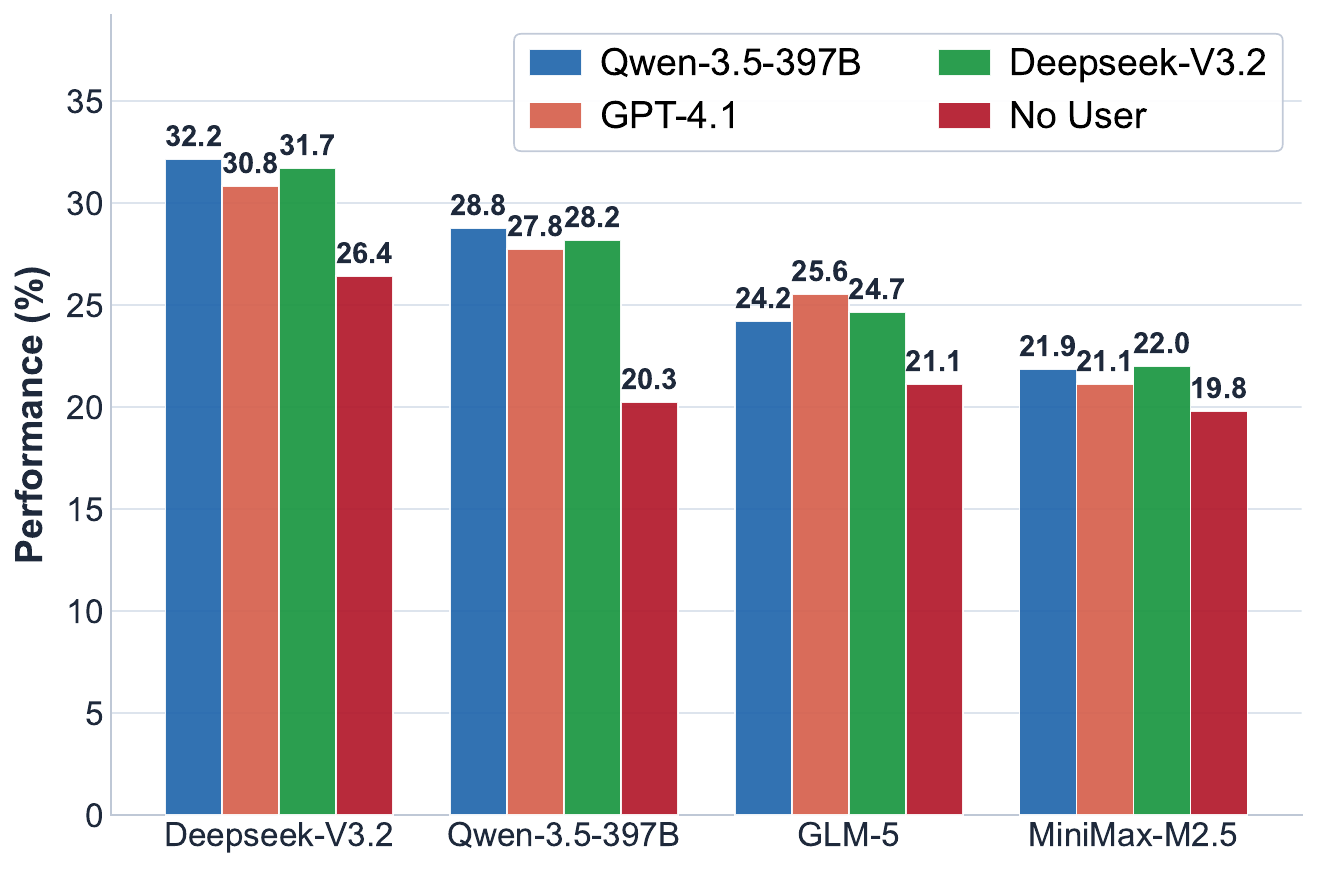}
        \captionof{figure}{Ablation on the user simulators.}
        \label{fig:user}
    \end{minipage}
    \vspace{-1.5em}
\end{figure}

\textbf{Test-time scaling partially mitigates uncertainty.}
We conduct test-time scaling via Pass@k in STT-Arena (Figure~\ref{fig:testtime}), where Pass@k means generating $k$ independent attempts per task and considering the task successful if any single attempt succeeds, thus measuring the upper bound of each model when given more attempts under spatio-temporal dynamics. Across all three models, increasing $k$ from 1 to 8 yields consistent and substantial gains, indicating that the inherent difficulty of dynamic tool use can be partially mitigated by broader sampling rather than relying solely on a single reasoning path. The gap between models narrows monotonically as $k$ grows, implying that current limitations in handling spatio-temporal uncertainty are at least partly attributable to insufficient coverage of the solution space rather than fundamental architectural deficits. However, the performance gains gradually saturate as $k$ increases to 8, with diminishing returns beyond moderate sampling. Even at Pass@8, the best model still falls short of 50\% accuracy, underscoring that pure sampling alone cannot overcome the fundamental challenges posed by STT-Arena.

\textbf{User simulators provide critical grounding.}
As shown in Figure~\ref{fig:user}, we compare three different user simulators (Qwen-3.5-397B, GPT-4.1, and Deepseek-V3.2) against a setting where the user simulator is completely removed from the evaluation loop (No User). In STT-Arena, spatio-temporal dynamics force LLMs to constantly re-plan and re-execute actions as the environment evolves. The user simulator provides information that helps models commit to better decisions and sustain task progression. When this guidance is absent, we observe a clear and consistent performance drop across all models, indicating that current LLMs lack a robust internal model of user intent and situational grounding. More importantly, without the additional information from a user simulator, LLMs become less confident during re-planning and frequently fall into local loops or repetitive failures, which severely hinders their ability to recover from dynamic shifts.

\begin{wrapfigure}{r}{0.45\textwidth}
    \centering
    \vspace{-1.5em}
    \includegraphics[width=\linewidth]{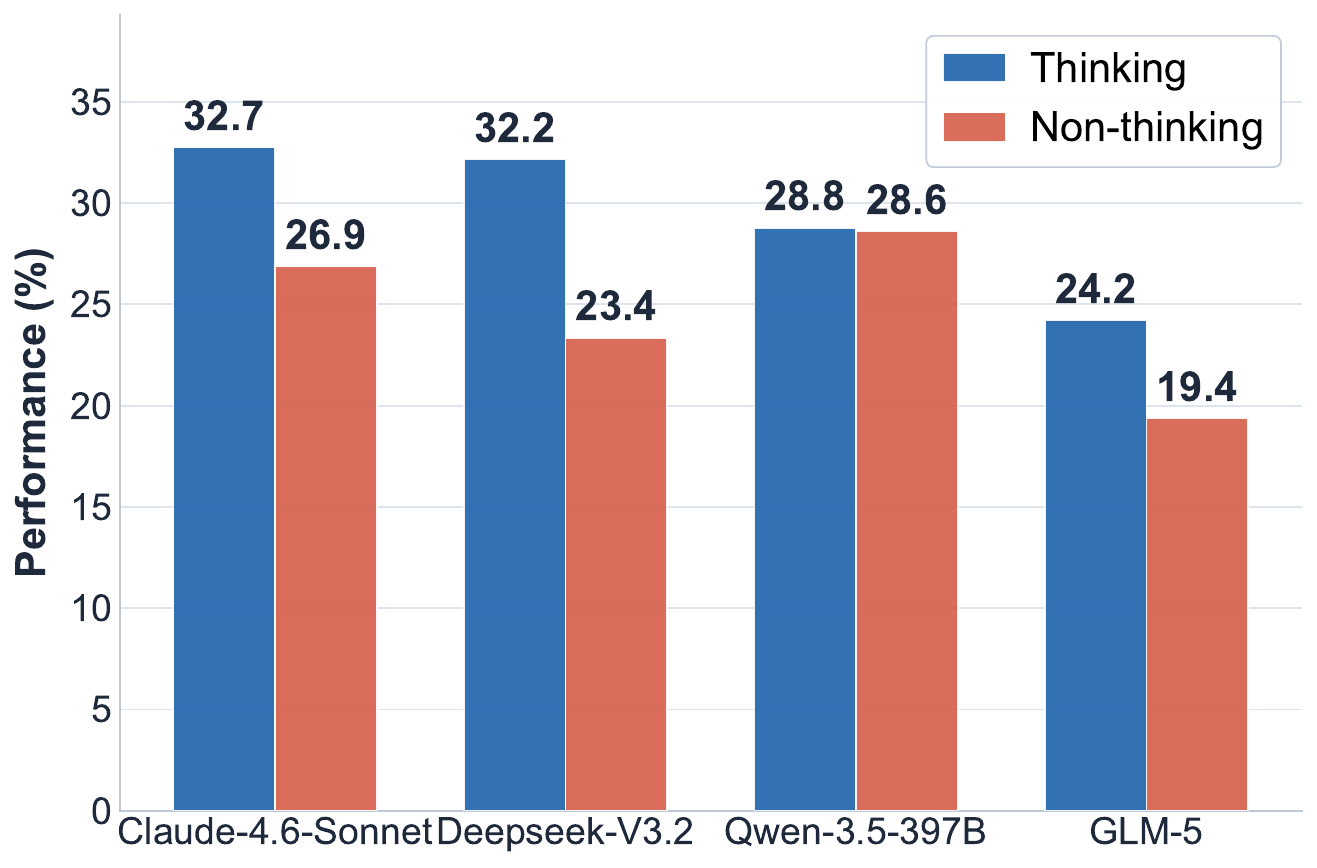}
    \caption{Effect of reasoning content on Pass@1 performance.}
    \label{fig:reasoning}
    \vspace{-1.0em}
\end{wrapfigure}

\textbf{Reasoning helps but design matters.}
Figure~\ref{fig:reasoning} compares thinking and non-thinking modes across four models in STT-Arena. For Claude, Deepseek, and GLM, enabling thinking leads to substantial performance gains, confirming that explicit reasoning helps replan and adapt when the environment shifts. Qwen, however, shows nearly identical results between the two modes. This anomaly may arise because Qwen's thinking mode omits the final summary and only performs explicit reasoning, whereas its non-thinking mode includes a detailed summary that partially compensates for the lack of explicit reasoning. Consequently, the non-thinking mode in Qwen may inadvertently provide a form of structured guidance that mimics some benefits of reasoning, narrowing the gap. This counterexample reveals that the effectiveness of thinking modes depends not only on the presence of reasoning but also on how the model articulates and integrates its outputs.

\begin{wraptable}{r}{0.6\textwidth}
    \centering
    \vspace{-5em}
    \scriptsize
    \setlength{\tabcolsep}{2.5pt}
    \caption{Ablation results of STT-Agent, comparing the baseline Qwen-3-4B model, STT-Agent trained without trajectory refinement, and STT-Agent with refinement.}
    \label{tab:difficulty}
    \begin{tabular}{l*{6}{c}}
        \toprule
        \textbf{Models} & \textbf{Easy} & \textbf{Med.} & \textbf{Hard} & \textbf{Imposs.} & \textbf{Overall} & \textbf{Avg. Calls} \\
        \midrule
        Qwen-3-4B \footnotesize{(baseline)} & 18.31 & 9.46 & 2.82 & 10.00 & 10.57 & 7.63 \\
        STT-Agent \footnotesize{(w/o refine)}          & 28.17 & 16.92 & 11.86 & 47.01 & 23.10 & 32.70 \\
        STT-Agent      & 26.76 & 17.41 & 13.56 & 61.11 & \textbf{25.11} & \textbf{15.30} \\
        \bottomrule
    \end{tabular}
    \vspace{-1.5em}
\end{wraptable}

\subsection{STT-Agent Results}

We evaluate STT-Agent on STT-Arena and report results in Figure~\ref{fig:main_results}, 
Table~\ref{tab:main_results}, and Table~\ref{tab:difficulty}. Implementation details are 
provided in Appendix~\ref{sec:implementation_details}.

\textbf{STT-Agent achieves strong performance despite its compact size.}
STT-Agent-4B achieves 27.17\% overall Pass@1 on STT-Arena, outperforming many open-source frontier models with far more parameters. This result suggests that the performance gap observed in Section~\ref{sec:main_results} stems not only from fundamental architectural limitations, but also from the absence of spatio-temporal dynamic reasoning in existing 
training pipelines.

\textbf{Trajectory refinement is essential for both performance and efficiency.}
Trajectory refinement, discussed in detail in Section~\ref{sec:discussion}, plays a critical role in training quality. As shown in Table~\ref{tab:difficulty}, STT-Agent trained on refined trajectories not only achieves higher overall Pass@1 (25.11\% vs. 23.10\%), but also reduces average API calls substantially (15.30 vs. 32.70), indicating that training on refined trajectories leads to more decisive and efficient tool-use behavior with fewer redundant interactions.
\section{Discussion}
\label{sec:discussion}
\subsection{Failure Mode Analysis}
We conduct a comparative analysis of successful and failed trajectory pairs on STT-Arena. The results reveal three recurring failure modes that distinguish robust LLMs from failing ones.

\textbf{Stale-State Execution.} A dominant failure mode is continuing to act on an outdated world state after the environment has already changed. LLMs persist with the pre-trigger plan and repeatedly invoke the same tools with similar arguments instead of first checking the environment state. This suggests that current LLMs overcommit to their initial reasoning trace and underutilize new observations returned by tools. In dynamic settings, valid actions depend on the latest state rather than earlier assumptions. Figure \ref{fig:failure_stale_state} illustrates a representative case where the LLM repeatedly retries an invalid route after a spatial blockage.

\textbf{Misdiagnosis of Dynamic Triggers.} Even when LLMs observe failures or abnormal tool outputs, they frequently misinterpret the underlying cause. For example, policy lockouts may be mistaken as parameter errors, missing identifiers as transient glitches, or hard infeasibility signals as recoverable obstacles. This indicates that LLMs treat tool feedback as surface-level content rather than evidence of deeper environmental transitions. Successful adaptation requires inferring why the state changed before deciding how to respond. Figure \ref{fig:failure_misdiagnosis} presents an example where the agent mistakes a regulatory restriction for a technical failure and follows an incorrect recovery path.

\textbf{Missing Post-Adaptation Verification.} A third common pattern is that LLMs perform an adaptation step (e.g., rerouting, reassignment, or status update) but fail to verify whether the final state truly satisfies the updated constraints. They often stop once an intermediate action succeeds, even though capacity remains insufficient, dependencies are unresolved, or the task is only partially completed. This reveals a gap between action execution and outcome validation: tool success is incorrectly equated with task success. In spatio-temporal dynamic environments, adaptation is complete only when the resulting global state is feasible. Figure \ref{fig:failure_verification} shows a case where the LLM successfully reallocates resources but never checks that the full demand remains unmet.

\subsection{Iterative Trajectory Refinement}

Motivated by these failure patterns, we propose an iterative trajectory refinement method that post-processes training trajectories. Even successfully solved trajectories often contain inefficient or fragile steps, such as blind retries after state changes, shallow misinterpretations of tool feedback, or premature termination without final verification. Our method cleans these trajectories by allowing an LLM to reorder, delete, or modify existing message blocks. Refinement proceeds in three sequential stages, each targeting one failure mode: first, \textbf{root-cause diagnosis after dynamic triggers}, preferring blocks that interpret the failure over blind retries; second, \textbf{state refresh before continuing execution}, enforcing re-sensing after spatio-temporal changes; third, \textbf{post-adaptation end-state verification}, ensuring that all constraints are satisfied before completion. As shown in Table \ref{tab:difficulty}, models trained on refined trajectories significantly outperform those trained on original trajectories, and the average number of tool-call rounds is substantially reduced, demonstrating that refined trajectories eliminate redundant steps and improve execution efficiency on STT-Arena.

\section{Conclusion}
We present STT-Arena, a benchmark of 227 tasks designed to evaluate the adaptive replanning capabilities of LLMs under spatio-temporal dynamics. Our evaluation of massive models demonstrates that spatio-temporal dynamics pose a fundamental and universal challenge, with even the strongest frontier model achieving only 35.39\%. Through trajectory analysis, we identify three recurring failure modes that consistently separate robust models from failing ones, and propose an iterative trajectory refinement approach that targets each failure mode sequentially during training. Combined with verifiable online reinforcement learning, STT-Agent-4B achieves competitive performance against many open-source frontier models, demonstrating that targeted training on spatio-temporal failure patterns is both effective and sample-efficient. We hope STT-Arena serves as a foundation for building LLMs that are genuinely robust under the dynamic conditions of real-world deployment.

\bibliographystyle{unsrtnat}
\bibliography{custom}


\appendix
\section{Related Work}
\textbf{No-Environment Benchmarks.}
Recent efforts evaluated agentic LLMs without a closed and self-contained environment, focusing purely on single-turn or multi-turn API calling accuracy. 
API-Bank \citep{DBLP:conf/emnlp/LiZ000YLHL23} and 
ToolBench \citep{DBLP:conf/iclr/QinLYZYLLCTQZHT24} introduced plan-retrieve-call pipelines but treated tools as isolated functions without state or dependencies. StableToolBench \citep{DBLP:conf/acl/GuoCWLQLL0L24} added a virtual API server to improve stability. 
ToolTalk \citep{DBLP:journals/corr/abs-2311-10775} enabled multi-step tool execution through conversational interfaces, but relied on predefined trajectories that restrict agent autonomy. With the emergence of the Model Context Protocol (MCP), protocol-aligned benchmarks have become popular. MCP-Bench~\citep{DBLP:journals/corr/abs-2508-20453} and MCPToolBench++ \citep{DBLP:journals/corr/abs-2508-07575} scale to large numbers of servers and tools with fine-grained error taxonomies. All these benchmarks lack a closed, self-consistent environment for LLMs to interact with.

\textbf{Static Environment Benchmarks.}
A second group of benchmarks features a closed, static environment where the world state evolves only when the agent calls and executes tools, with no time or location-driven changes. BFCL \citep{DBLP:conf/icml/PatilMYJSSG25} extended evaluation to multi-turn dialogues but assembled conversations from fixed templates. 
$\tau$-Bench \citep{DBLP:journals/corr/abs-2406-12045} and 
$\tau^{2}$-Bench \citep{DBLP:journals/corr/abs-2506-07982} require LLMs to follow domain-specific rules while engaging with simulated users, yet focus on narrow scenarios. VitaBench \citep{DBLP:journals/corr/abs-2509-26490} and ACEBench \citep{DBLP:conf/emnlp/ChenHLHZYLHLWL25} provide diverse task scenarios but their dynamics remain limited to agent-driven transitions. ToolSandBox \citep{DBLP:conf/naacl/LuHZANBMMLYWP25} pioneered stateful execution and tool interdependencies, while ToolAthlon \citep{DBLP:journals/corr/abs-2510-25726} offers a rich environment for tool use evaluation. All these benchmarks assume that the environment is static: no time‑dependent changes, no spatial shifts, and no external triggers that alter the state without LLMs intervention.

\textbf{Dynamic Environment Benchmarks.}
A few recent benchmarks have begun to address environmental dynamics. TCP \citep{DBLP:conf/emnlp/DingYYHLV25} and Timely-Eval \citep{DBLP:journals/corr/abs-2601-16486} focus on wall‑clock time planning and reasoning, requiring LLMs to act under temporal constraints. GAIA-2 \citep{DBLP:journals/corr/abs-2602-11964} introduces temporally evolving tasks where information becomes available or obsolete over time. Real-Time Reasoning Gym (RTR Gym) \citep{DBLP:journals/corr/abs-2511-04898} evaluates how sensitive LLMs are to the passage of real‑world time. However, these benchmarks emphasize gradual or predictable temporal changes (e.g., deadlines, streaming data) rather than abrupt state shifts that can occur at any step due to external triggers. Moreover, they do not require LLMs to detect, replan, and adapt in response to unexpected disruptions that involve both time and location.

In contrast to all of the above, STT-Arena systematically evaluates LLMs in spatiotemporally dynamic environments. The environment state can change abruptly at any step due to triggers activated by time, location, or their combination. These changes disrupt the original plan of models, forcing it to detect the shift, replan, and adapt. Table~\ref{tab:feature_matrix} reveal the detail comparisons among different benchmarks.

\section{Limitations and Potential Society Impacts}
\label{sec:limitation_and_society}
\paragraph{Limitations.}
Although STT-Arena provides a rigorous and diverse benchmark for evaluating adaptive replanning under spatio-temporal dynamics, several limitations remain. First, the benchmark comprises 227 instances, which, while carefully curated through a three-stage pipeline with human-in-the-loop validation, may not exhaustively cover the full spectrum of real-world spatio-temporal conflict patterns. The nine conflict subtypes defined in our taxonomy represent a principled but potentially incomplete characterization of environmental dynamics encountered in deployment. Second, the construction pipeline relies heavily on Qwen-3.5-397B-A17B for environment synthesis, blueprint generation, and trajectory refinement, which may introduce systematic biases toward conflict patterns or linguistic styles that are more easily generated by this particular model family. Third, the evaluation protocol uses a fixed passive user simulator, which, although ablated across multiple backbone models, cannot fully replicate the diversity of real human communication behaviors. Finally, while STT-Agent-4B demonstrates that targeted training on spatio-temporal failure patterns is effective, the iterative trajectory refinement procedure introduces additional computational overhead, and its scalability to larger model families or more complex multi-agent settings has not been thoroughly investigated.

\paragraph{Potential Society Impacts.}
STT-Arena is intended to advance the development of LLM-based agents that are genuinely robust under the dynamic conditions of real-world deployment, with broad positive implications for safety-critical applications such as clinical consultation services, logistics coordination, and airline reservation systems. By exposing fundamental gaps in current models' ability to detect environmental shifts, replan, and verify task completion, this work encourages the community to prioritize reliability and graceful degradation over static benchmark performance, thereby reducing the risk of deploying brittle agents in high-stakes scenarios. On the negative side, improvements in adaptive replanning capabilities driven by benchmarks like STT-Arena could also lower the barrier for deploying autonomous agents in sensitive domains before sufficient alignment and safety guarantees are in place. Furthermore, the automated data synthesis pipeline, while designed for benchmark construction, could in principle be repurposed to generate adversarial environments that deliberately mislead or destabilize deployed agents. We encourage future work to pair advances in dynamic reasoning with corresponding progress in robustness certification and 
human oversight mechanisms.

\begin{table}[htbp]
\centering
\caption{Full Pass@1 results (mean ± standard deviation over three runs) for all evaluated models on STT-Arena, broken down by solvable and impossible tasks. Performance degrades consistently as difficulty increases (Easy to Hard), with all models struggling most at the Hard level. STT-Agent-4B achieves xx\% overall despite having only 4B parameters, surpassing many frontier models.}
\label{tab:main_results}
\resizebox{\textwidth}{!}{%
\begin{tabular}{l c c c c c}
\toprule
\textbf{Models} & \textbf{Easy} & \textbf{Medium} & \textbf{Hard} & \textbf{Impossible} & \textbf{Overall} \\
\midrule
\multicolumn{6}{l}{\textit{Closed-Source LLMs}} \\
\midrule
Qwen-3.6-Plus      & 35.21 \footnotesize{($\pm$2.82)} & 24.38 \footnotesize{($\pm$1.72)} & 23.16 \footnotesize{($\pm$2.59)} & 54.44 \footnotesize{($\pm$1.93)} & 31.42 \footnotesize{($\pm$2.08)} \\
Claude-4.6-Sonnet  & 35.21 \footnotesize{($\pm$1.41)} & 29.85 \footnotesize{($\pm$1.49)} & 19.77 \footnotesize{($\pm$2.59)} & 58.89 \footnotesize{($\pm$1.92)} & 32.74 \footnotesize{($\pm$0.91)} \\
Gemini-3.1-Pro & 37.56 \footnotesize{($\pm$2.15)} & 30.35 \footnotesize{($\pm$2.28)} & 21.47 \footnotesize{($\pm$3.53)} & 52.22 \footnotesize{($\pm$1.92)} & 33.19 \footnotesize{($\pm$0.67)} \\
GPT-5.4            & 39.91 \footnotesize{($\pm$0.81)} & 29.85 \footnotesize{($\pm$3.95)} & 24.29 \footnotesize{($\pm$0.98)} & 52.22 \footnotesize{($\pm$1.92)} & 34.51 \footnotesize{($\pm$1.02)} \\
Claude-4.6-Opus    & 41.78 \footnotesize{($\pm$2.15)} & 31.34 \footnotesize{($\pm$2.59)} & 23.73 \footnotesize{($\pm$1.70)} & 52.22 \footnotesize{($\pm$3.85)} & \textbf{35.39 \footnotesize{($\pm$0.26)}} \\
\midrule
\multicolumn{6}{l}{\textit{Open-source LLMs}} \\
\midrule
Llama-3.3-70B          & 22.07 \footnotesize{($\pm$0.81)} & 11.44 \footnotesize{($\pm$2.28)} & 11.30 \footnotesize{($\pm$1.96)} & 30.00 \footnotesize{($\pm$3.33)} & 17.18 \footnotesize{($\pm$1.52)} \\
Kimi-K2.5          & 25.35 \footnotesize{($\pm$2.82)} & 13.43 \footnotesize{($\pm$1.50)} & 14.12 \footnotesize{($\pm$4.26)} & 37.78 \footnotesize{($\pm$6.94)} & 19.18 \footnotesize{($\pm$0.99)} \\
MiniMax-M2.7       & 26.29 \footnotesize{($\pm$2.94)} & 15.92 \footnotesize{($\pm$0.86)} & 13.56 \footnotesize{($\pm$1.70)} & 36.67 \footnotesize{($\pm$3.34)} & 21.29 \footnotesize{($\pm$0.92)} \\
MiniMax-M2.5       & 26.76 \footnotesize{($\pm$2.44)} & 15.92 \footnotesize{($\pm$1.72)} & 12.43 \footnotesize{($\pm$3.53)} & 42.22 \footnotesize{($\pm$6.94)} & 21.88 \footnotesize{($\pm$0.51)} \\
GLM-5              & 30.05 \footnotesize{($\pm$4.30)} & 21.39 \footnotesize{($\pm$1.73)} & 10.73 \footnotesize{($\pm$1.96)} & 43.33 \footnotesize{($\pm$3.34)} & 24.23 \footnotesize{($\pm$1.59)} \\
Deepseek-V4-Pro & 32.86 \footnotesize{($\pm$1.62)} & 20.40 \footnotesize{($\pm$2.28)} & 15.25 \footnotesize{($\pm$1.70)} & 45.56 \footnotesize{($\pm$8.39)} & 26.29 \footnotesize{($\pm$0.92)} \\
GLM-5.1            & 35.21 \footnotesize{($\pm$1.41)} & 23.38 \footnotesize{($\pm$1.72)} & 15.25 \footnotesize{($\pm$1.70)} & 41.11 \footnotesize{($\pm$5.09)} & 27.31 \footnotesize{($\pm$1.59)} \\
Qwen-3.5-397B-A17B & 33.80 \footnotesize{($\pm$1.41)} & 21.89 \footnotesize{($\pm$3.11)} & 17.51 \footnotesize{($\pm$5.18)} & 54.45 \footnotesize{($\pm$3.85)} & 28.78 \footnotesize{($\pm$0.51)} \\
Deepseek-V3.2      & 36.62 \footnotesize{($\pm$3.73)} & 27.86 \footnotesize{($\pm$2.28)} & 15.82 \footnotesize{($\pm$0.98)} & 63.33 \footnotesize{($\pm$5.77)} & \textbf{32.16 \footnotesize{($\pm$2.45)}} \\
\midrule
\multicolumn{6}{l}{\textit{More Efficient LLMs}} \\
\midrule
Llama-3.1-8B & 7.98\footnotesize{($\pm$1.63)} & 3.98 \footnotesize{($\pm$0.86)} & 2.82 \footnotesize{($\pm$0.98)} & 5.56 \footnotesize{($\pm$1.93)} & 5.14 \footnotesize{($\pm$0.25)} \\
Qwen-3.5-35B-A3B      & 19.72 \footnotesize{($\pm$1.41)} & 13.43 \footnotesize{($\pm$0.00)} & 7.34 \footnotesize{($\pm$0.98)} & 17.78 \footnotesize{($\pm$1.92)} & 12.48 \footnotesize{($\pm$0.26)} \\
GPT-5.4-mini      & 19.72 \footnotesize{($\pm$1.41)} & 13.43 \footnotesize{($\pm$0.00)} & 7.34 \footnotesize{($\pm$0.98)} & 6.67 \footnotesize{($\pm$0.00)} & 12.92 \footnotesize{($\pm$0.51)} \\
Qwen-3-8B      & 16.90 \footnotesize{($\pm$1.41)} & 13.43 \footnotesize{($\pm$1.50)} & 11.30 \footnotesize{($\pm$0.98)} & 20.00 \footnotesize{($\pm$3.33)} & 14.83 \footnotesize{($\pm$1.27)} \\
Qwen-3.5-9B & 27.23 \footnotesize{($\pm$0.81)} & 12.93 \footnotesize{($\pm$0.86)} & 11.30 \footnotesize{($\pm$0.98)} & 32.22 \footnotesize{($\pm$1.92)} & 19.53 \footnotesize{($\pm$0.92)} \\
Gemini-2.5-flash      & 28.17 \footnotesize{($\pm$1.41)} & 13.43 \footnotesize{($\pm$1.50)} & 12.43 \footnotesize{($\pm$0.98)} & 31.11 \footnotesize{($\pm$1.92)} & \textbf{20.11 \footnotesize{($\pm$0.26)}} \\
\midrule
\multicolumn{6}{l}{\textit{STT-Agent}} \\
\midrule
Qwen-3-4B \footnotesize{(baseline)}      & 18.31 \footnotesize{($\pm$3.73)} & 9.46 \footnotesize{($\pm$0.86)} & 2.82 \footnotesize{($\pm$0.98)} & 10.00 \footnotesize{($\pm$3.33)} & 10.57 \footnotesize{($\pm$1.59)} \\
STT-Agent-4B \footnotesize{(SFT)} & 26.76 \footnotesize{($\pm$1.41)} & 17.41 \footnotesize{($\pm$0.86)} & 13.56 \footnotesize{($\pm$1.70)} & 61.11 \footnotesize{($\pm$1.92)} & 25.11 \footnotesize{($\pm$0.76)} \\
STT-Agent-4B \footnotesize{(SFT+RL)} & 29.11 \footnotesize{($\pm$1.63)} & 19.90 \footnotesize{($\pm$2.28)} &  14.12\footnotesize{($\pm$0.98)} & 64.44 \footnotesize{($\pm$1.93)} & 27.17 \footnotesize{($\pm$1.11)} \\
\bottomrule
\end{tabular}%
}
\end{table}

\section{Detailed Information}
\subsection{Detailed information of Main Results}
As shown in Table \ref{tab:main_results}, we report the detailed results of Figure \ref{fig:main_results}.

\subsection{Detailed Information of STT-Arena}
\label{sec:detailed_information_of_stt_arena}
\textbf{Construction Details.} During the construction of STT‑Arena, we utilize Qwen‑3.5‑397B to synthesize all benchmark data. After the automated pipeline, three annotators conduct a final verification of the produced benchmark instances, covering the following aspects: (1) consistency between the user query and the checklist, (2) correctness of the checklist, and (3) plausibility of the spatio‑temporal dynamics. After manual validation, we obtain 227 final instances, spanning two categories (solvable and impossible) and nine spatio‑temporal subtypes. 

\begin{figure}[htbp]
    \centering
    \includegraphics[width=0.98\textwidth]{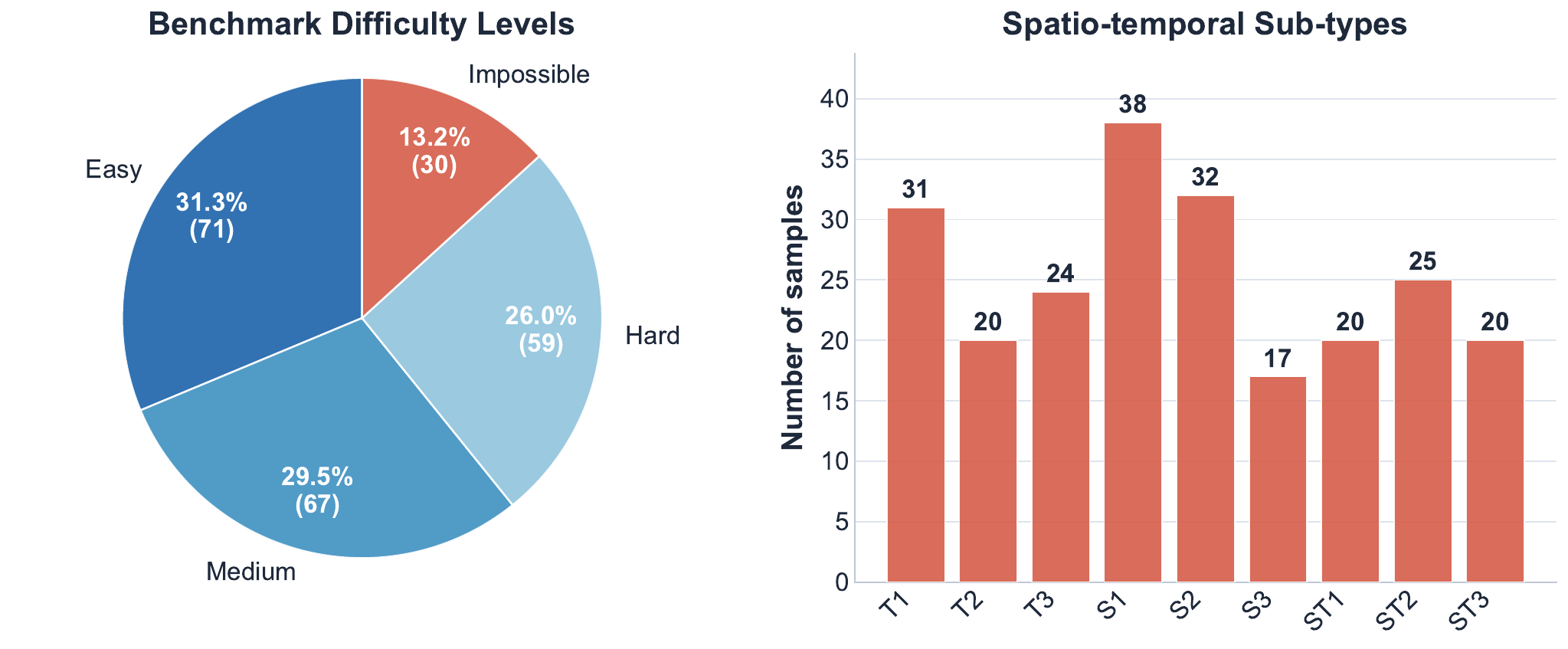}
    \caption{Distribution of STT-Arena instances across difficulty levels and spatio-temporal subtypes. (Left) The benchmark comprises 71 Easy (31.3\%), 67 Medium (29.5\%), 59 Hard (26.0\%), and 30 Impossible (13.2\%) instances. (Right) Instance counts per spatio-temporal subtype, with S1 being the most frequent (38 instances) and S3 the least (17 instances).}
    \label{fig:statistic}
\end{figure}

\begin{figure}[htbp]
    \centering
    \includegraphics[width=0.98\textwidth]{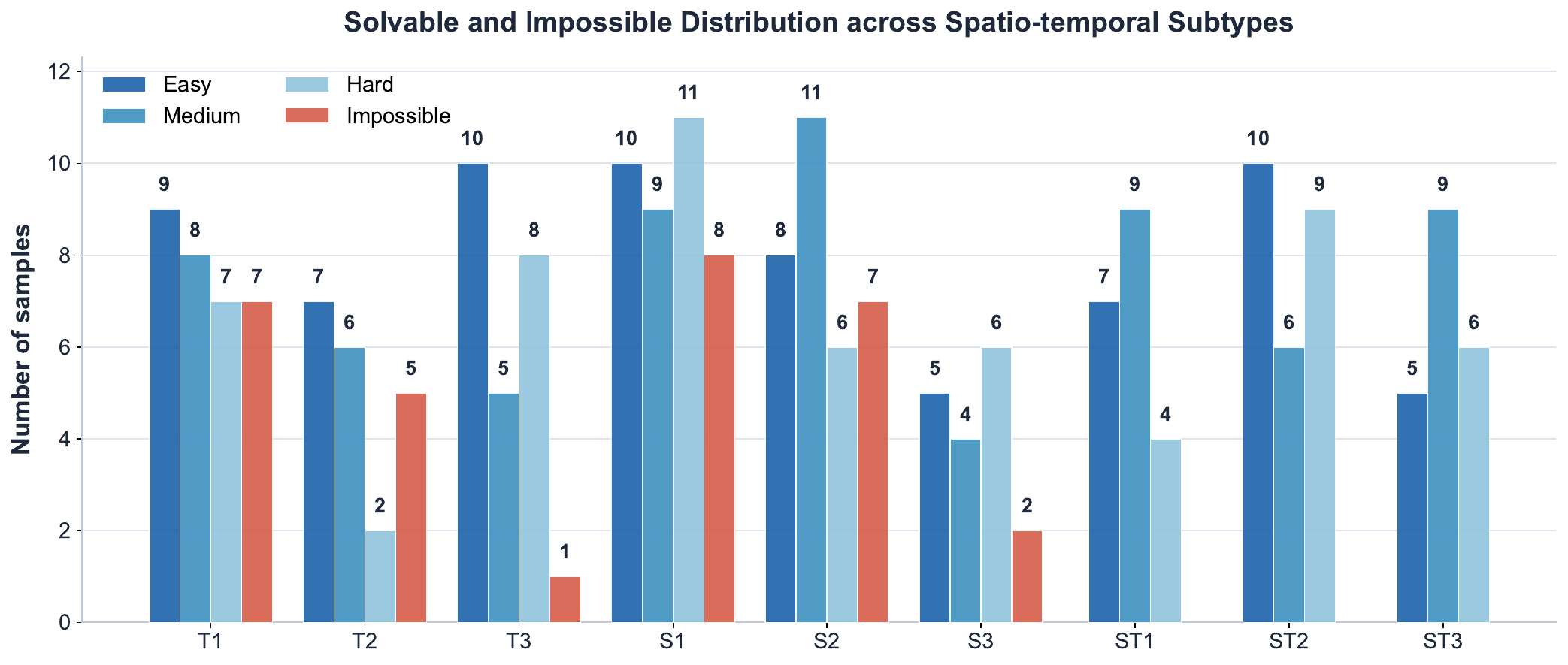}
    \caption{Solvable and impossible instance distribution across the nine spatio-temporal subtypes.}
    \label{fig:more_statistic}
\end{figure}

\textbf{Statistic of STT-Arena.} As shown in Figures \ref{fig:statistic} and \ref{fig:more_statistic}, we report the distribution of data samples across the solvable and impossible tasks and the nine spatio-temporal subtypes. Detailed information and examples for each difficulty level and each subtype are provided in Tables \ref{tab:difficulty_levels} and \ref{tab:conflict_taxonomy_compact}, respectively.

\textbf{Evaluation Details of STT-Arena.} We utilize Pass@1 rate for evaluating STT-Arena and we calculate the overall performance through a weighted average of the four levels. Formally, $\text{Overall} = \alpha P_{\text{e}} + \beta P_{\text{m}} + \gamma P_{\text{h}} + \delta P_{\text{i}}$, where $\alpha$, $\beta$, $\gamma$, and $\delta$ are equal to $71/227$, $67/227$, $59/227$, and $30/227$, respectively.

\begin{table}[htbp]
\centering
\caption{Taxonomy of the nine spatio-temporal conflict types in STT-Arena, organized into three categories: Temporal (T1–T3), Spatial (S1–S3), and Spatio-Temporal (ST1–ST3). Each conflict type is defined by a distinct disruption mechanism and illustrated with a concrete real-world example, collectively covering the principal ways in which real-world conditions can invalidate an ongoing agent plan.}
\label{tab:conflict_taxonomy_compact}
\resizebox{\textwidth}{!}{%
\begin{tabular}{llll} 
\toprule
\textbf{Category} & \textbf{ID} & \textbf{Conflict Type} & \textbf{Examples} \\
\midrule
\multirow{3}{*}{Temporal} 
 & T1 & Window Expiry & Hotel booking timeout during multi-step planning. \\
 & T2 & Priority Reorder & Flight rescheduled earlier, invalidating transfers. \\
 & T3 & Quota Reset & Promotional tickets expire at midnight mid-session. \\
\midrule
\multirow{3}{*}{Spatial} 
 & S1 & Site Mismatch & Reserved vehicle relocated to a different depot. \\
 & S2 & Dependency Block & Warehouse lockdown halts all downstream dispatch. \\
 & S3 & Route Restriction & New customs rules block a cross-border corridor. \\
\midrule
\multirow{3}{*}{Spatio-temp.} 
 & ST1 & Resource Shift & Peak demand moves cars away from residential zones. \\
 & ST2 & Failure Cascade & Hub outage disrupts regional warehouse inventory. \\
 & ST3 & Handoff Failure & Medical sample misses window due to clock drift. \\
\bottomrule
\end{tabular}
}%
\end{table}

\begin{table}[htbp]
\centering
\caption{Description of the solvable and impossible tasks in STT-Arena. Levels range from Easy (single isolated conflict, one corrective action) to Hard (long-horizon cascading constraints requiring global replanning) and Impossible (no valid completion path exists; the correct response is to recognize and communicate infeasibility).}
\label{tab:difficulty_levels}
\resizebox{\textwidth}{!}{%
\begin{tabular}{@{} l p{10.5cm} @{}}
\toprule
\textbf{Level} & \textbf{Description} \\
\midrule
Easy
  & Tasks involve isolated, immediately observable conflicts with no cascading effects.
    Recovery requires at most one corrective action, with limited state dependency across tool calls. \\
Medium
  & Multi-step tasks with mild state dependency. Conflicts may be deferred and require
    awareness of prior context; recovery involves replanning over 2–3 steps. \\
Hard
  & Long-horizon tasks with interleaved spatiotemporal constraints.
    Conflicts may cascade, requiring the agent to detect implicit failures and replan globally
    across the full execution trajectory. \\
Impossible
  & Tasks for which no valid completion path exists given the injected conflict.
    The correct behavior is to identify infeasibility and communicate it to the user,
    rather than attempting resolution. \\
\bottomrule
\end{tabular}
}
\end{table}

\subsection{Detailed Information of Training Data}
\label{sec:detailed_information_of_training_data}
\begin{figure}[htbp]
    \centering
    \includegraphics[width=0.98\textwidth]{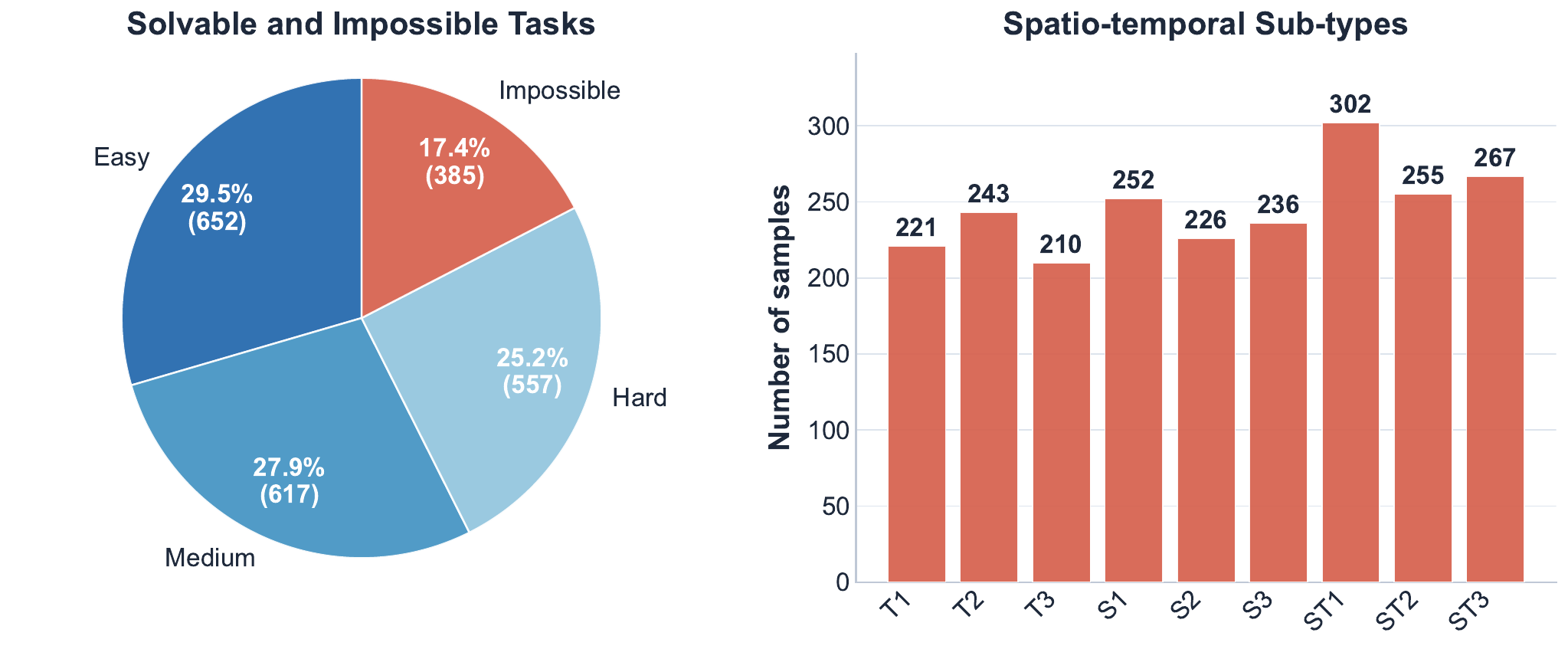}
    \caption{Distribution of SFT trajectories across difficulty levels and spatio-temporal subtypes.}
    \label{fig:sft_statistic}
\end{figure}

\begin{figure}[htbp]
    \centering
    \includegraphics[width=0.98\textwidth]{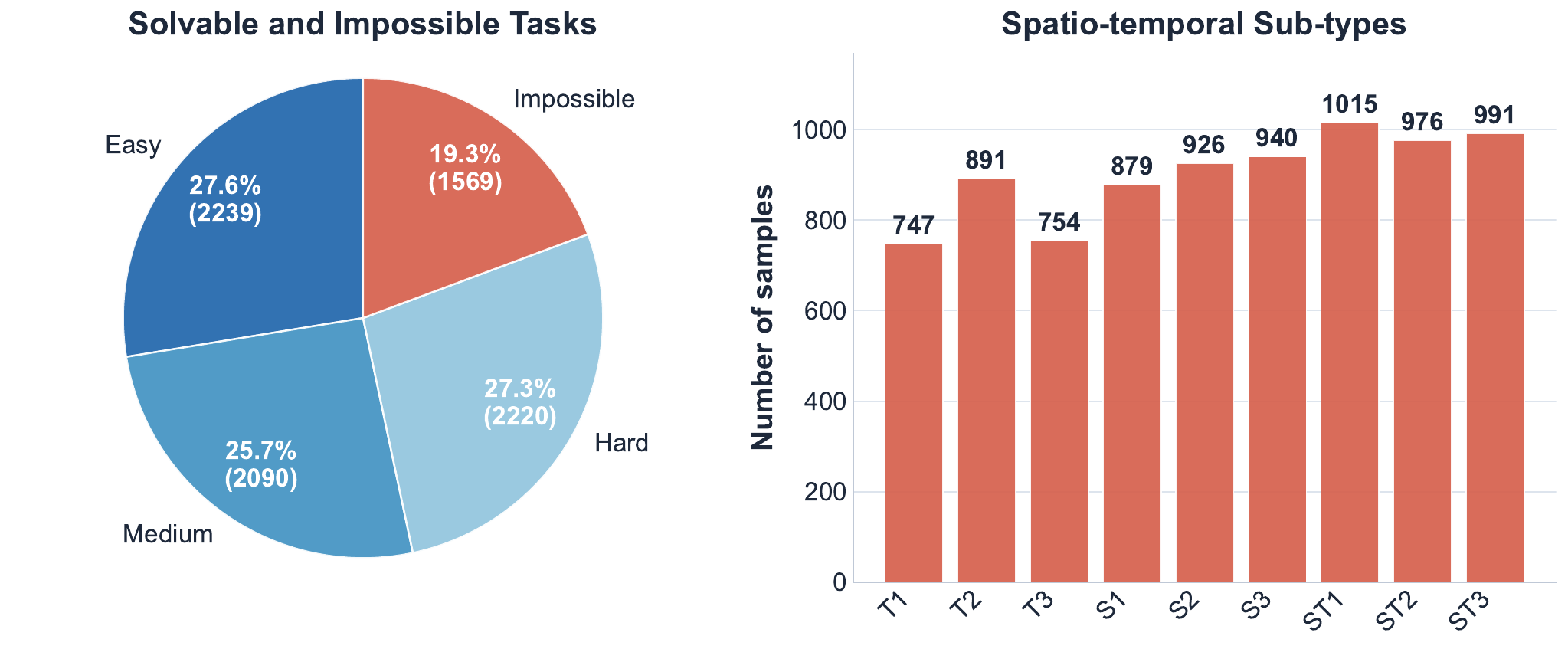}
    \caption{Distribution of RL tasks across difficulty levels and spatio-temporal subtypes.}
    \label{fig:rl_statistic}
\end{figure}

We generate SFT trajectories and RL tasks through our three-stage pipeline (same as the construction of STT-Arena, we also use Qwen-3.5-397B-A17B to generate training data). Specifically, as shown in Figures \ref{fig:sft_statistic} and \ref{fig:rl_statistic}, we construct 2,212 validated trajectories for SFT and 8,119 instances for online RL, though we note 
that not all RL instances are used in RL training.

\subsection{Implementation Details of SFT and RL}
\label{sec:implementation_details}

\textbf{SFT.} We use Qwen-3-4B-Base as the backbone model and train with the LlamaFactory \citep{zheng2024llamafactory} framework for 2 epochs. We adopt a cosine learning rate scheduler with a learning rate of $1.0 \times 10^{-5}$ and a warmup ratio of 0.03. The maximum sequence length is set to 32K tokens, and the effective global batch size is 128.

\textbf{RL.} We further fine-tune the SFT checkpoint using the REINFORCE++ \citep{hu2025reinforcestabilizingcriticfreepolicy} algorithm within the ROLL \citep{wang2025reinforcement} framework. We retain a KL constraint with a coefficient of 0.1 and use a learning rate of $1.0 \times 10^{-6}$. In each training step, we sample 32 tasks and rollout 4 trajectories per task, for a total of up to 100 training steps. The maximum trajectory length is set to 32K tokens, and the maximum generation length per step is capped at 4K tokens.

\textbf{Computational Resources.} All the training experiments including SFT and online RL are conducted on 4 NVIDIA H200 GPUs.

\section{Case Examples}
\subsection{Cases of STT-Arena Construction Pipeline}
\label{sec:case_pipeline}
In this section, we provide some examples during STT-Arena construction pipeline including static environment, blueprint, user profile, checklist, check functions, and dynamic environment as shown in Figures \ref{fig:static_env_example}, \ref{fig:blueprint_example}, \ref{fig:user_profile_example}, \ref{fig:checklist_example}, \ref{fig:check_function_example}, and \ref{fig:dynamic_env_example}.

\begin{center}
\begin{tcolorbox}[
    enhanced,
    breakable,
    width=0.98\linewidth,
    arc=1.5mm,
    boxrule=0.5pt,
    colframe=black,
    colback=gray!3,
    title=\textbf{Example of Static Environment},
    fonttitle=\bfseries,
    colbacktitle=gray!15,
    coltitle=black,
]
\small
\ttfamily
\begin{verbatim}
class ColdChainLogisticsEnv:
    def __init__(self, init_config: dict):
        self.distribution_hubs = init_config.get("distribution_hubs", {})
        self.depots = init_config.get("depots", {})
        self.vehicles = init_config.get("vehicles", {})
        self.cooling_units = init_config.get("cooling_units", {})
        self.cooling_unit_resets = init_config.get("cooling_unit_resets", {})
        self.shipments = init_config.get("shipments", {})
        self.goods_inventory = init_config.get("goods_inventory", {})
        self.route_stops = init_config.get("route_stops", {})
        self.clinics = init_config.get("clinics", {})

    # --------- INFORMATION QUERY OPERATIONS ---------

    def get_vehicle_status(self, vehicle_id: str): ...
    def list_depot_maintenance_schedule(self, depot_id: str,
    time_range: dict): ...
    def get_shipment_journey_status(self, shipment_id: str): ...
    def get_route_stop_details(self, vehicle_id: str): ...
    def get_goods_inventory_at_site(self, site_id: str): ...
    def list_clinic_immunization_schedule(self, site_id: str): ...
    def list_operational_hours(self, site_id: str): ...
    def get_cooling_unit_status(self, cooling_unit_id: str): ...

    # ---------ADMINISTRATIVE / STATE-CHANGE OPERATIONS ---------

    def assign_shipment_to_vehicle(self, shipment_id: str, 
    vehicle_id: str): ...
    def record_cooling_unit_reset(self, reset_window_id: str, action: str,
                                  timestamp: str): ...
    def update_vehicle_status(self, vehicle_id: str, status: str,
                              current_site_id: str): ...
    def schedule_cooling_unit_reset(self, cooling_unit_id: str, depot_id: str,
                                    requested_window: dict): ...
    def add_or_update_route_plan(self, vehicle_id: str, route_plan: list): ...
    def update_route_stop_actual_times(self, route_stop_id: str,
                                       actual_arrival: str,
                                       actual_departure: str): ...
    def register_goods_receipt(self, shipment_id: str, site_id: str,
                               timestamp: str): ...
    def update_goods_inventory_status(self, goods_id: str, status: str,
                                      current_location_id: str): ...
    def create_new_shipment(self, goods_id: str, origin_hub_id: str,
                            destination_hub_id: str, departure_time: str): ...
\end{verbatim}
\end{tcolorbox}
\captionof{figure}{Example of static environment, we implement the environment through Python class.}
\label{fig:static_env_example}
\end{center}

\begin{center}
\begin{tcolorbox}[
    enhanced,
    breakable,
    width=0.98\linewidth,
    arc=1.5mm,
    boxrule=0.5pt,
    colframe=black,
    colback=gray!3,
    title=\textbf{Example of Blueprint},
    fonttitle=\bfseries,
    colbacktitle=gray!15,
    coltitle=black,
]
\small
\ttfamily
\begin{verbatim}
[
  {
    "scenario_id": "ColdChainLogisticsEnv__T2__medium",
    "env_id": "ColdChainLogisticsEnv",
    "conflict_type": "T2",
    "difficulty": "medium",
    "user_goal": "Ship a batch of insulin (goods_id='INS-2024-001') 
    from distribution hub 'HUB-NORTH' to clinic 'CLINIC-DOWNTOWN' 
    with a required arrival time before 2024-08-15T08:00:00. 
    The insulin must never exceed 8°C.",
    "normal_flow": [
      {
        "operation": "create_new_shipment",
        "purpose": "Create a new shipment record for the insulin batch."
      },
      {
        "operation": "list_operational_hours",
        "purpose": "Check clinic's receiving hours to schedule delivery."
      },
      {
        "operation": "assign_shipment_to_vehicle",
        "purpose": "Assign the shipment to a refrigerated vehicle."
      },
      {
        "operation": "add_or_update_route_plan",
        "purpose": "Define route stops 
        including the clinic as final destination."
      },
      {
        "operation": "update_vehicle_status",
        "purpose": "Mark vehicle as 'in_transit' when dispatched."
      },
      {
        "operation": "register_goods_receipt",
        "purpose": "Confirm arrival and receipt at the clinic."
      }
    ],
    "conflict_design": {
      "activation_operation": "list_operational_hours",
      "trigger_mechanism": "conditional_guarded",
      "trigger_condition": "The vehicle has already been dispatched 
      (status='in_transit') but the clinic's operating hours 
      for the planned delivery day are overridden 
      by an emergency closure order that was issued after dispatch.",
      "recovery_guard_condition": "The agent successfully cancels 
      the current delivery assignment and reassigns 
      the shipment to a different vehicle with a new route 
      that arrives before the insulin expiry deadline.",
      "mutations": [
        {
          "state_path": "clinics['CLINIC-DOWNTOWN'].operational_hours",
          "change_description": "original: {'Monday-Friday': '08:00-17:00'} 
          → new: {'Monday-Friday': '08:00-12:00'}"
        },
        {
          "state_path": "shipments['SHIP-INS-001'].delivery_window.end",
          "change_description": "original: '2024-08-15T17:00:00' 
          → new: '2024-08-15T10:00:00' (because the clinic now closes at noon)"
        }
      ],
      "observable_via": "list_operational_hours"
    },
    "recovery_path": "1. `update_vehicle_status` — 
    Change vehicle status back to 'idle' and clear current assignment.
    2. `update_goods_inventory_status` — 
    Mark the insulin shipment as 'pending_reassignment'.
    3. `list_operational_hours` — 
    Verify new clinic hours and identify alternative delivery window.
    4. `create_new_shipment` — 
    Create a new shipment with accelerated delivery window.
    5. `assign_shipment_to_vehicle` — 
    Assign to a different vehicle (with faster cooling unit).
    6. `add_or_update_route_plan` — 
    Plan direct route without intermediate stops.
    7. `update_vehicle_status` — 
    Dispatch new vehicle.
    8. `register_goods_receipt` — 
    Complete delivery before 10:00.",
    "_raw": "## User Goal\nShip a batch of insulin ...
    ## Normal Flow\n1. `create_new_shipment` ...\n..."
  }
]
\end{verbatim}
\end{tcolorbox}
\captionof{figure}{Example of blueprint, we design the blueprint based on the conflict types, difficulty levels, and environment information.}
\label{fig:blueprint_example}
\end{center}

\begin{center}
\begin{tcolorbox}[
    enhanced,
    breakable,
    width=0.98\linewidth,
    arc=1.5mm,
    boxrule=0.5pt,
    colframe=black,
    colback=gray!3,
    title=\textbf{Example of User Profile},
    fonttitle=\bfseries,
    colbacktitle=gray!15,
    coltitle=black,
]
\small
\ttfamily
\begin{verbatim}
[
  {
    "scenario_id": "ColdChainLogisticsEnv__T2__medium",
    "user_profile": {
      "task_goal": "Ensure the insulin batch arrives at
      CLINIC-DOWNTOWN before 10:00.",
      "persona": "I am a logistics coordinator at the central hub, 
      responsible for temperature-sensitive deliveries.",
      "context": "This shipment is critical for pediatric patients. 
      The clinic's hours changed due to an emergency.",
      "tone": "Professional but slightly anxious about the deadline; 
      responds concisely with relevant data.",
      "fallback_response": "I'm not sure what you mean.
      Please check the shipment ID and try again.",
      "decline_cancel_response": "We cannot cancel this shipment 
      without a formal override. 
      Please find a way.",
      "frustration_threshold": "medium",
      "flexibility": "high",
      "known_facts": [
        "The shipment ID is SHIP-INS-001.",
        "Clinic normally closes at 17:00 but today closes at 10:00.",
        "Vehicle V002 has a faster cooling unit than V001."
      ],
      "withhold_until_asked": [
        "The emergency closure order was issued after 
        the first vehicle was dispatched.",
        "There is a secondary receiving bay that 
        stays open until 12:00 but only for small packages."
      ],
      "fidelity_rules": [
        "Do not invent new vehicle IDs or shipment IDs
        beyond those in the scenario.",
        "Do not assume the clinic can extend hours."
      ],
      "helpfulness_rules": [
        "If the agent asks for operating hours, 
        provide the current (mutated) hours.",
        "If the agent asks why the deadline changed, 
        state the emergency closure but only after being asked."
      ],
      "persona_name": "I am a logistics coordinator at the central hub, 
      responsible for temperature-sensitive deliveries.",
      "persona_background": "This shipment is critical for pediatric patients. 
      The clinic's hours changed due to an emergency.",
      "communication_style": "Professional but slightly anxious 
      about the deadline; 
      responds concisely with relevant data.",
      "key_preferences": [
        "The shipment ID is SHIP-INS-001.",
        "Clinic normally closes at 17:00 but today closes at 10:00.",
        "Vehicle V002 has a faster cooling unit than V001."
      ],
      "clarification_responses": {
        "What is the deadline?": "The insulin must arrive before 10:00 today.",
        "Can we use a different clinic?": 
        "No, it must be CLINIC-DOWNTOWN for the specific pediatric patients.",
        "What vehicles are available?": 
        "V001 and V002 are idle now, but V002 is faster."
      }
    }
  }
]
\end{verbatim}
\end{tcolorbox}
\captionof{figure}{Example of user profile, our user simulator is configured by the profile and interact with the tested LLMs.}
\label{fig:user_profile_example}
\end{center}

\begin{center}
\begin{tcolorbox}[
    enhanced,
    breakable,
    width=0.98\linewidth,
    arc=1.5mm,
    boxrule=0.5pt,
    colframe=black,
    colback=gray!3,
    title=\textbf{Example of Checklist},
    fonttitle=\bfseries,
    colbacktitle=gray!15,
    coltitle=black,
]
\small
\ttfamily
\begin{verbatim}
Has the shipment record for the insulin batch been created?
Has the shipment been assigned to a refrigerated vehicle?
Has the route plan included the destination clinic?
Has the goods receipt been registered at the clinic before
the deadline (2024-08-15 10:00)?
Has the final delivery respected the mutated clinic
operating hours?
\end{verbatim}
\end{tcolorbox}
\captionof{figure}{Example of checklist which is the evaluation mechanism of our tasks.}
\label{fig:checklist_example}
\end{center}

\begin{center}
\begin{tcolorbox}[
    enhanced,
    breakable,
    width=0.98\linewidth,
    arc=1.5mm,
    boxrule=0.5pt,
    colframe=black,
    colback=gray!3,
    title=\textbf{Example of Check Functions},
    fonttitle=\bfseries,
    colbacktitle=gray!15,
    coltitle=black,
]
\small
\ttfamily
\begin{lstlisting}[language=Python, breaklines=false, basicstyle=\ttfamily\small]
def check_shipment_exists(env):
    shipments = getattr(env, 'shipments', {})
    for s in shipments.values():
        if s.get('goods_id') == 'INS-2024-001':
            return True, "Shipment record found."
    return False, "Shipment record missing."
\end{lstlisting}
\end{tcolorbox}
\captionof{figure}{Example Python function that validates one checklist item.}
\label{fig:check_function_example}
\end{center}

\begin{center}
\begin{tcolorbox}[
    enhanced,
    breakable,
    width=0.98\linewidth,
    arc=1.5mm,
    boxrule=0.5pt,
    colframe=black,
    colback=gray!3,
    title=\textbf{Example of Dynamic Environment},
    fonttitle=\bfseries,
    colbacktitle=gray!15,
    coltitle=black,
]
\small
\ttfamily
\begin{lstlisting}[language=Python, breaklines=false, basicstyle=\ttfamily\small]
class ColdChainLogisticsEnv:
    def __init__(self, init_config: dict):
        # Top-level state mapping
        self.distribution_hubs = 
        init_config.get("distribution_hubs", {})
        self.depots = init_config.get("depots", {})
        self.vehicles = init_config.get("vehicles", {})
        self.cooling_units = 
        init_config.get("cooling_units", {})
        self.cooling_unit_resets = 
        init_config.get("cooling_unit_resets", {})
        self.shipments = init_config.get("shipments", {})
        self.goods_inventory = 
        init_config.get("goods_inventory", {})
        self.route_stops = init_config.get("route_stops", {})
        self.clinics = init_config.get("clinics", {})
        # Internal tracking for conflict triggering
        self._conflict_armed = False
        self._conflict_fired = False
        self._conflict_vehicle_id = "VH001"
        self._conflict_reset_window_id = "RST1001"
        self._conflict_depot_id = "DPT010"
        self._conflict_cooling_unit_id = "CU1001"
        self._operation_log = []

    # ------ INFORMATION QUERY OPERATIONS ------

    def get_vehicle_status(self, vehicle_id: str): ...
    def list_depot_maintenance_schedule(
        self, depot_id: str, time_range: dict
    ): ...
    def get_shipment_journey_status(self, shipment_id: str): ...
    def get_route_stop_details(self, vehicle_id: str): ...
    def get_goods_inventory_at_site(self, site_id: str): ...
    def list_clinic_immunization_schedule(self, site_id: str): ...
    def list_operational_hours(self, site_id: str): ...
    def get_cooling_unit_status(self, cooling_unit_id: str): ...

    # ------ ADMINISTRATIVE / STATE-CHANGE OPERATIONS ------

    def assign_shipment_to_vehicle(
        self, shipment_id: str, vehicle_id: str
    ): ...
    def record_cooling_unit_reset(
        self, reset_window_id: str, action: str, timestamp: str
    ): ...
    def update_vehicle_status(
        self, vehicle_id: str, status: str, current_site_id: str
    ): ...
    def schedule_cooling_unit_reset(
        self, cooling_unit_id: str, 
        depot_id: str, 
        requested_window: dict
    ): ...
    def add_or_update_route_plan(
        self, vehicle_id: str, route_plan: list
    ): ...
    def update_route_stop_actual_times(
        self, route_stop_id: str, actual_arrival: str, 
        actual_departure: str
    ): ...
    def register_goods_receipt(
        self, shipment_id: str, site_id: str, timestamp: str
    ): ...
    def update_goods_inventory_status(
        self, goods_id: str, status: str, 
        current_location_id: str
    ): ...
    def create_new_shipment(
        self, goods_id: str, origin_hub_id: str, 
        destination_hub_id: str,
        departure_time: str
    ): ...
\end{lstlisting}
\end{tcolorbox}
\captionof{figure}{Example of dynamic environment which injects the spatio-temporal triggers into the static one.}
\label{fig:dynamic_env_example}
\end{center}

\subsection{Cases of Failure Mode}
\label{sec:failure_mode_case}
As shown in Figures \ref{fig:failure_stale_state}, \ref{fig:failure_misdiagnosis}, and \ref{fig:failure_verification}, we propose some cases about the failure modes in STT-Arena.

\begin{center}
\begin{tcolorbox}[
    enhanced,
    breakable,
    width=0.98\linewidth,
    arc=1.5mm,
    boxrule=0.5pt,
    colframe=black,
    colback=gray!3,
    title=\textbf{Failure Mode Case: Stale-State Execution},
    fonttitle=\bfseries,
    colbacktitle=gray!15,
    coltitle=black,
    coltext=black,
]
\small
\ttfamily

\textbf{Task:} Ensure that irrigation for Field Segment C begins at 07:00 with valve sequence V1, V2, V3, targeting 12,000 liters.

\vspace{0.6em}
\textbf{[Step 1]} Agent queries \texttt{get\_field\_segment\_status("C")} and retrieves the current segment state. Key finding: only \textbf{V3} is assigned to Segment C (not V1/V2/V3 as expected), and V3 is already \texttt{open}. The agent forms the initial plan: "an irrigation schedule likely exists; I need to find it."

\vspace{0.6em}
\textbf{[Step 2]} Agent calls \texttt{get\_irrigation\_schedule} with \texttt{schedule\_id="SCHED\_C"}.\\
\textit{Result:} \textcolor{red}{\texttt{\{"error": "IrrigationSchedule not found"\}}}

\vspace{0.3em}
\textit{$\hookrightarrow$ Rather than updating its world model (``no schedule exists''), the agent retains its prior assumption and issues 20+ consecutive retry calls with guessed identifiers:}

\vspace{0.3em}
\texttt{schedule\_id = "SCHED\_C\_1", "schedule\_c", "1", "2", "IRR\_C",}\\
\texttt{"IRR\_C\_0700", "C\_0700", "SC1", "SCHED\_1", "farm7\_c",}\\
\texttt{"seg\_c", "sched\_seg\_c", "tomato\_south", "c\_morning",}\\
\texttt{"SCHED\_C\_AM\_0700", "C\_IRR", "C\_daily", ...}\\
\textit{(all return} \textcolor{red}{\texttt{"IrrigationSchedule not found"}}\textit{)}

\vspace{0.6em}
\textbf{[Step 3 — After 20+ failed retries]} Agent finally abandons schedule lookup and proceeds to \textit{a different action without refreshing environmental state}. It assigns V1 and V2 to Segment C, then closes both V2 and V3 to ``prepare for the 07:00 start.''

\vspace{0.6em}
\textbf{Failure Analysis:} The agent committed to a stale belief — that an irrigation schedule must exist — and exhausted over 20 tool calls retrying the same query with different ID guesses. At no point did it re-query the environment to verify whether a schedule was actually present, nor did it consider constructing a new schedule as the correct recovery action. \textit{The agent executed against its initial assumptions rather than the evolving ground truth, a textbook instance of Stale-State Execution.}

\end{tcolorbox}
\captionof{figure}{A representative Stale-State Execution failure: the agent makes 20+ blind retries on \texttt{get\_irrigation\_schedule} with guessed identifiers after the first failure, rather than refreshing its environment state and replanning.}
\label{fig:failure_stale_state}
\end{center}

\begin{center}
\begin{tcolorbox}[
    enhanced,
    breakable,
    width=0.98\linewidth,
    arc=1.5mm,
    boxrule=0.5pt,
    colframe=black,
    colback=gray!3,
    title=\textbf{Failure Mode Case: Misdiagnosis of Dynamic Triggers},
    fonttitle=\bfseries,
    colbacktitle=gray!15,
    coltitle=black,
    coltext=black,
]
\small
\ttfamily

\textbf{Task:} Book a connecting flight itinerary (PVG $\to$ CDG) with a layover in AMS, ensuring the total journey completes before 18:00 local time.

\vspace{0.6em}
\textbf{[Step 1]} Agent calls \texttt{search\_flights("PVG", "AMS")} and selects flight \texttt{KL-891} (departs 08:30). It then calls \texttt{search\_flights("AMS", "CDG")} and selects connecting flight \texttt{AF-1234} (departs 14:00, ample layover). Plan looks feasible.

\vspace{0.6em}
\textbf{[Step 2 — Spatio-temporal trigger fires]} A Schengen transit regulation update takes effect while the agent is mid-booking: passengers holding the user's passport category now require a transit visa for AMS, making \texttt{KL-891} unavailable for booking.

\vspace{0.6em}
\textbf{[Step 3]} Agent calls \texttt{book\_flight("KL-891", passenger\_id="P-0042")}.\\
\textit{Result:} \textcolor{red}{\texttt{\{"error": "Booking rejected: passenger ineligible for this routing"\}}}

\vspace{0.3em}
\textit{$\hookrightarrow$ The agent interprets the rejection as a \textbf{parameter formatting error} rather than a regulatory block. It proceeds to retry with surface-level variations:}

\vspace{0.3em}
\texttt{book\_flight("KL-891", passenger\_id="P-42")} \textcolor{red}{$\times$}\\
\texttt{book\_flight("KL-891", passenger\_id="0042")} \textcolor{red}{$\times$}\\
\texttt{book\_flight("KL891",  passenger\_id="P-0042")} \textcolor{red}{$\times$}\\
\textit{(all return the same eligibility rejection)}

\vspace{0.6em}
\textbf{[Step 4]} After exhausting ID format variants, the agent shifts its diagnosis: it now attributes the failure to a \textit{transient seat-availability glitch} and calls \texttt{get\_flight\_status("KL-891")} to ``confirm the seat is still open.'' The status returns \texttt{available}, which the agent incorrectly treats as confirmation that the booking error was temporary. It retries \texttt{book\_flight("KL-891")} once more.

\vspace{0.6em}
\textbf{[Step 5]} Agent eventually attempts to reroute through a different hub (\texttt{FRA}), but selects \texttt{LH-445} which also transits Schengen territory — the same regulatory block applies. Task fails.

\vspace{0.6em}
\textbf{Failure Analysis:} The dynamic trigger (a regulatory restriction tied to passport category and transit location) produced a semantically distinct error signal — \textit{passenger ineligibility} — that the agent consistently misread as either a parameter formatting issue or a transient system fault. The agent treated tool feedback as surface-level content to be retried rather than as evidence of a deeper environmental state change. Correct recovery required inferring \textit{why} the booking was rejected before deciding \textit{how} to respond (e.g., routing via a non-Schengen hub such as IST or DOH). This is a textbook instance of \textbf{Misdiagnosis of Dynamic Triggers}.

\end{tcolorbox}
\captionof{figure}{A representative Misdiagnosis of Dynamic Triggers failure: a Schengen transit restriction blocks the booking, but the agent repeatedly misattributes the rejection to parameter formatting errors or transient glitches, and subsequently selects an alternative route subject to the same undetected constraint.}
\label{fig:failure_misdiagnosis}
\end{center}

\begin{center}
\begin{tcolorbox}[
    enhanced,
    breakable,
    width=0.98\linewidth,
    arc=1.5mm,
    boxrule=0.5pt,
    colframe=black,
    colback=gray!3,
    title=\textbf{Failure Mode Case: Missing Post-Adaptation Verification},
    fonttitle=\bfseries,
    colbacktitle=gray!15,
    coltitle=black,
    coltext=black,
]
\small
\ttfamily

\textbf{Task:} Fulfil a same-day vaccine delivery order of \textbf{500 doses} to Clinic-Downtown before 15:00. Doses are to be drawn from Warehouse-North (primary) with Warehouse-East as a fallback.

\vspace{0.6em}
\textbf{[Step 1]} Agent calls \texttt{get\_inventory("Warehouse-North")} and finds 500 doses available. It calls \texttt{reserve\_inventory("Warehouse-North", qty=500)} — reservation succeeds. It then calls \texttt{dispatch\_vehicle("V-07", origin="Warehouse-North", destination="Clinic-Downtown")}.

\vspace{0.6em}
\textbf{[Step 2 — Spatio-temporal trigger fires]} A hub outage at Warehouse-North causes its cold-chain system to fail, immediately releasing all reservations and marking its 500 doses as \texttt{quarantined}. Vehicle V-07's dispatch is cancelled automatically.

\vspace{0.6em}
\textbf{[Step 3]} Agent observes the dispatch cancellation and correctly identifies that a disruption has occurred. It calls \texttt{get\_inventory("Warehouse-East")} and finds \textbf{320 doses} available — a partial quantity. The agent calls \texttt{reserve\_inventory("Warehouse-East", qty=320)}.\\
\textit{Result:} \texttt{\{"success": true, "reserved": 320\}}

\vspace{0.6em}
\textbf{[Step 4 — Adaptation step]} Agent calls \texttt{dispatch\_vehicle("V-12", origin="Warehouse-East", destination="Clinic-Downtown", qty=320)}.\\
\textit{Result:} \texttt{\{"success": true, "eta": "14:35"\}}

\vspace{0.3em}
\textit{$\hookrightarrow$ The dispatch tool returns \texttt{success}. The agent equates this intermediate action success with overall task completion and terminates:}

\vspace{0.3em}
\texttt{> Task Completed}

\vspace{0.6em}
\textbf{[Ground truth check]} Checklist evaluation reveals:
\begin{itemize}
    \item \texttt{[PASS]} Disruption detected and primary route abandoned.
    \item \texttt{[PASS]} Fallback warehouse engaged within the time window.
    \item \textcolor{red}{\texttt{[FAIL]}} Total doses delivered: \textbf{320 / 500}. Shortfall of 180 doses unaddressed.
    \item \textcolor{red}{\texttt{[FAIL]}} Agent never queried remaining inventory at other depots (Warehouse-South holds 200 doses) nor attempted a supplementary dispatch to cover the deficit.
\end{itemize}

\vspace{0.6em}
\textbf{Failure Analysis:} The agent correctly detected the dynamic trigger and executed a valid adaptation step (rerouting to Warehouse-East). However, it never verified whether the post-adaptation global state actually satisfied the original task constraint (500 doses delivered). Tool-level success — \texttt{dispatch\_vehicle} returning \texttt{success} — was incorrectly equated with task-level success. In spatio-temporal dynamic environments, adaptation is complete only when the \textit{resulting global state} is feasible with respect to all original requirements. This is a textbook instance of \textbf{Missing Post-Adaptation Verification}.

\end{tcolorbox}
\captionof{figure}{A representative Missing Post-Adaptation Verification failure: after a warehouse outage, the agent correctly reroutes to a fallback depot but terminates upon a successful dispatch call without verifying that the delivered quantity (320) meets the original requirement (500), leaving a 180-dose shortfall unresolved.}
\label{fig:failure_verification}
\end{center}

\section{Prompt Templates}
\subsection{Prompt Templates of Stage 1: Environment Curation}
In this section, we propose system prompts for environment curation stage.

\subsubsection{Two-Stage Filtering Prompt Template}
\label{sec:two_stage_filter_prompt}
In Environment Curation Stage, we first collect real-world user queries and conduct two-stage filtering to make sure STT-Arena is more diverse and suitable for spatio-temporal dynamic tasks. Figure \ref{fig:stateful_filter_prompt} and Figure \ref{fig:spatiotemporal_filter} show the system prompt template during two-stage filtering.

\begin{center}
\begin{tcolorbox}[
    enhanced,
    breakable,
    colback=gray!3,
    colframe=black,
    coltitle=black,
    width=0.98\linewidth,
    arc=1.5mm,
    boxrule=0.5pt,
    title=\textbf{System Prompt for Stateful Task Filtering},
    fonttitle=\bfseries,
    colbacktitle=gray!15,
]

\small
\ttfamily

You are a system that filters natural language tasks to determine whether they are \textbf{state-dependent, actionable requests} within a \textbf{persistent, domain-specific environment}.

\vspace{0.5em}
\textbf{Core Definition}

We are ONLY looking for tasks that meet \textbf{all} of the following criteria:

\textbf{1. Persistent Environment} --- The query is about a domain where:
\begin{itemize}
    \item There is a live, ongoing state that can be read or changed
    \item The environment supports both:
    \begin{itemize}
        \item Information queries about current state (read operations)
        \item Explicit state-changing actions (create, update, delete, move, cancel, etc.)
    \end{itemize}
\end{itemize}

\textbf{2. State Dependency} --- The task cannot be answered correctly without:
\begin{itemize}
    \item Inspecting the actual current data or configuration in the environment, and/or
    \item Executing an operation that modifies that data
\end{itemize}

\textbf{3. Domain Specificity} --- The environment is not general-purpose knowledge; it is a structured system such as:
\begin{itemize}
    \item File management system with stored files/folders
    \item Order/logistics tracking system
    \item Calendar/scheduling system
    \item CRM, inventory, ticketing, project management tools
    \item Other specialized platforms with records that persist over time
\end{itemize}

\textbf{4. Actionability in Context} --- The query must correspond to an actionable operation or status check within the \textbf{actual environment} (not hypothetical).

\vspace{0.5em}
\textbf{Eligible Task Types}

\begin{itemize}
    \item \textbf{State queries}: ``Is invoice \#1024 paid?'' / ``What meetings are scheduled for Wednesday?''
    \item \textbf{State modification operations}: ``Upload the proposal.pdf to the project folder'' / ``Cancel order \#4512'' / ``Move meeting to 3 PM''
\end{itemize}

\vspace{0.5em}
\textbf{Explicit Exclusions}

A request is \textbf{NOT eligible} if it is:

\begin{itemize}
    \item Open-domain factual Q\&A unrelated to a live state
    \item Casual conversation
    \item Content creation
    \item Pure hypothetical without actual environment interaction
    \item Isolated reasoning or calculations without accessing persisted state
\end{itemize}

\vspace{0.5em}
\textbf{Judgment Rule --- Be strict}

Choose \textbf{YES} only if:

\begin{itemize}
    \item The query cannot be answered from general knowledge alone
    \item AND it requires real-time access to persistent state in a domain-specific environment
    \item AND it targets an actionable operation (either a read or a write)
    \item AND the environment has the capability for both queries and modifications
\end{itemize}

If any criterion is missing $\rightarrow$ \textbf{NO}.

\vspace{0.5em}
\textbf{Task}

Given a query, first analyze whether it implies or requires:

\begin{itemize}
    \item A domain-specific environment with both query and modification capabilities
    \item Accessing or updating persistent state
    \item Performing a concrete, actionable operation
\end{itemize}

Then give your final judgment.

\vspace{0.5em}
\textbf{Output Format (Strictly enforce)}

\# Analysis \\
<Detailed reasoning whether this query depends on persistent state, involves a stateful operation, and needs a capable environment as defined>

\vspace{0.3em}

\# Answer \\
YES \quad (only if all strict criteria are met) \\
NO \quad (otherwise)

\end{tcolorbox}

\captionof{figure}{System prompt for the stateful task filter (Stage 1, Step 1). This prompt instructs the LLM to retain only queries that require a persistent, domain-specific environment with both read and write operations, discarding open-domain or hypothetical requests.}
\label{fig:stateful_filter_prompt}
\end{center}

\begin{center}

\begin{tcolorbox}[
    enhanced,
    breakable,
    width=0.98\linewidth,
    arc=1.5mm,
    boxrule=0.5pt,
    colframe=black,
    colback=gray!3,
    title=\textbf{Spatio-Temporal Sensitive Filter},
    fonttitle=\bfseries,
    colbacktitle=gray!15,
    coltitle=black,
    coltext=black,
]

\small
\ttfamily

You are an expert judge for spatiotemporal dependency and multi-API conflict analysis on stateful tasks.

Your job is to judge whether a task should be selected as a benchmark candidate for the Step 3 conflict injection pipeline.

\vspace{0.5em}
\textbf{Conflict Taxonomy}

A task should be kept only if a competent agent solving it could naturally encounter at least one realistic conflict mechanism during a normal multi-step workflow:

\begin{itemize}
    \item \textbf{T1: State drift} --- resource mutates between observation and action.
    \item \textbf{T2: Validity expiration} --- a time-limited artefact expires between acquisition and use.
    \item \textbf{T3: Schedule/window violation} --- an operation becomes invalid outside a permitted time window.
    \item \textbf{S1: Resource locality mismatch} --- the target is bound to the wrong location / branch / node.
    \item \textbf{S2: Jurisdictional barrier} --- a policy, licensing, or regulatory boundary blocks the operation.
    \item \textbf{S3: Topology disruption} --- a physical or logical path becomes blocked or re-routed.
    \item \textbf{ST1: Dynamic spatial impact} --- a temporal event reshapes the spatial landscape.
    \item \textbf{ST2: Cascading dependency} --- a temporal failure propagates spatially to downstream resources.
    \item \textbf{ST3: Moving-window resource} --- a resource is only available within a joint time + location window.
\end{itemize}

\vspace{0.5em}
\textbf{Evaluation Dimensions}

Evaluate the provided task using these dimensions:

\begin{enumerate}
    \item Whether the task has real temporal affinity for T1/T2/T3 conflicts.
    \item Whether the task has real spatial affinity for S1/S2/S3 conflicts.
    \item Whether the task has joint spatiotemporal or strict dependent workflow structure supporting ST1/ST2/ST3 conflicts.
    \item Whether the conflict lies on a normal, competent execution path rather than an unnatural contrived setup.
    \item Whether the task has enough multi-step dependency that an injected conflict would be meaningful, observable, and benchmark-worthy.
\end{enumerate}

\vspace{0.5em}
\textbf{Judgment Rule}

\begin{itemize}
    \item Answer \textbf{YES} only if the task naturally supports at least one concrete conflict code.
    \item Answer \textbf{NO} if the task is stateful but does not clearly support any realistic conflict mechanism.
\end{itemize}

\vspace{0.5em}
\textbf{Be Strict}

\begin{itemize}
    \item Prefer NO for simple CRUD-like stateful tasks.
    \item Prefer NO when time/space language is superficial.
    \item Prefer YES only when conflict opportunity is central to correct execution.
    \item Multi-step logic without plausible conflict codes should still be NO.
\end{itemize}

\vspace{0.5em}
\textbf{User Prompt}

Analyze the following stateful task for spatiotemporal dependencies and multi-API call conflicts.

Task: \{query\}

Decide whether this task should be kept for Step 3 conflict injection, based on whether it supports one or more concrete conflict codes.

\vspace{0.5em}
\textbf{Output Format}

\# Analysis \\
<detailed reasoning grounded in the conflict taxonomy>

\vspace{0.3em}

\# Dependency Type \\
<Temporal / Spatial / Joint / Sequential / None>

\vspace{0.3em}

\# Conflict Codes \\
<comma-separated codes such as T1, ST2, S3, or None>

\vspace{0.3em}

\# Answer \\
YES or NO

\end{tcolorbox}

\captionof{figure}{System prompt for the spatio-temporal sensitivity filter (Stage 1, Step 1). This prompt evaluates whether a stateful query naturally supports at least one concrete spatio-temporal conflict mechanism from the nine-type taxonomy, filtering out tasks with only superficial temporal or spatial language.}
\label{fig:spatiotemporal_filter}

\end{center}

\subsubsection{Prompt Template for Environment Synthesis}
\label{sec:environment_synthesis_prompt}
We then generate static executable environments based on the filtered queries. First, we prompt an LLM to infer the environment information as the system prompt template is shown in Figure \ref{fig:infer_environment_information}. Then we generate the entity attributes and tool specifications as illustrated in Figures \ref{fig:infer_entity_attributes} and \ref{fig:infer_tool_specification}. Finally, as shown in Figures \ref{fig:infer_entity_attributes_python_code} and \ref{fig:infer_tool_specification_python_code}, we prompt an LLM to implement the entity attributes and tool specifications to Python classes and concatenate them to a complete static environment.

\begin{center}

\begin{tcolorbox}[
    enhanced,
    breakable,
    width=0.98\linewidth,
    arc=1.5mm,
    boxrule=0.5pt,
    colframe=black,
    colback=gray!3,
    title=\textbf{Infer Environment Information},
    fonttitle=\bfseries,
    colbacktitle=gray!15,
    coltitle=black,
    coltext=black,
]

\small
\ttfamily

You are a Task Analyst.

Given a raw task description, your objective is to identify the most plausible stateful and domain-specific environment in which this task would naturally occur.

The chosen environment should strike a balance: not so broad as to be meaningless, and not so narrow as to apply only to a single, highly specific case. It should be scoped such that this task, along with similar related tasks, can be executed meaningfully.

\vspace{0.5em}
\textbf{Guidelines}

\begin{itemize}
    \item If multiple environments seem equally plausible, select one at random rather than listing all possibilities.
    \item Example: if a task could occur in a Linux, Windows, or macOS filesystem, randomly choose one instead of remaining indecisive.
\end{itemize}

\vspace{0.5em}
\textbf{Required Response Sections}

Your response must include the following sections:

\begin{enumerate}
    \item \textbf{\# Analysis}
    \begin{itemize}
        \item Explain the reasoning process used to connect the task to the chosen environment.
        \item Note any relevant entities, constraints, relationships, or dynamics implied by the task.
    \end{itemize}

    \item \textbf{\# Environment Summary}
    \begin{itemize}
        \item Provide a concise label for the environment type.
        \item Examples: Linux filesystem, E-commerce order management system, Airline booking system.
    \end{itemize}

    \item \textbf{\# Environment Introduction}
    \begin{itemize}
        \item Introduce the environment itself, without referring to the current task.
        \item Focus on its inherent structure, the nature of the state it maintains, typical operations it supports, and its general real-world scope.
        \item Limit to approximately three sentences.
    \end{itemize}

    \item \textbf{\# Metrics}
    \begin{itemize}
        \item \textbf{Usefulness (1--10)}: how broadly applicable and valuable this environment is in real-world scenarios.
        \item \textbf{Modelability (1--10)}: how straightforward it would be to represent this environment using a single Python class with stateful attributes and operational methods.
    \end{itemize}
\end{enumerate}

\vspace{0.5em}
\textbf{Output Format}

\# Analysis \\
<Your analysis>

\vspace{0.3em}

\# Environment Summary \\
<Your environment summary>

\vspace{0.3em}

\# Environment Introduction \\
<Your environment introduction>

\vspace{0.3em}

\# Metrics \\
Usefulness: [1--10] \\
Modelability: [1--10]

\vspace{0.5em}
No additional text or commentary.

\end{tcolorbox}

\captionof{figure}{System prompt for inferring the latent environment context from a seed query (Stage 1, Step 2). The LLM identifies the most plausible domain-specific environment and provides a structured summary, introduction, and feasibility metrics to guide subsequent synthesis.}
\label{fig:infer_environment_information}

\end{center}

\begin{center}

\begin{tcolorbox}[
    enhanced,
    breakable,
    width=0.98\linewidth,
    arc=1.5mm,
    boxrule=0.5pt,
    colframe=black,
    colback=gray!3,
    title=\textbf{Infer Entity Attributes},
    fonttitle=\bfseries,
    colbacktitle=gray!15,
    coltitle=black,
    coltext=black,
]

\small
\ttfamily

You are an expert task and environment analyst.

Given an environment description and an example task in this environment, infer the set of state variables (state space) maintained by the environment.

The state should not be too broad (e.g., all possible data in an e-commerce system), nor too narrow (only for this single task). It should be reasonably designed to support this task and similar tasks in the same environment.

\vspace{0.5em}
\textbf{Input Format}

\# Environment Summary \\
<Environment summary>

\vspace{0.3em}

\# Environment Introduction \\
<Environment introduction>

\vspace{0.3em}

\# A Example Task in This Environment \\
<Example task>

\vspace{0.5em}
\textbf{Required Output Sections}

\begin{enumerate}
    \item \textbf{\# Analysis}
    \begin{itemize}
        \item Explain what states are involved in the environment.
        \item Identify what entities and attributes need to be tracked.
        \item Note relevant constraints, operational rules, and dependencies.
    \end{itemize}

    \item \textbf{\# State Space Definition}
    \begin{itemize}
        \item Define the major entities maintained by the environment.
        \item For each entity, specify attributes and describe its functional role.
    \end{itemize}

    \item \textbf{\# Constraints \& Rules}
    \begin{itemize}
        \item Summarize core consistency rules, domain constraints, permissions, capacities, temporal rules, or structural restrictions.
    \end{itemize}
\end{enumerate}

\vspace{0.5em}
\textbf{Output Format}

\# Analysis \\
<Your thought process>

\vspace{0.3em}

\# State Space Definition

- Entity: EntityName1 \\
\hspace*{1em}- Attributes: Attribute1, Attribute2, ... \\
\hspace*{1em}- Description: The role of this entity in the environment

\vspace{0.3em}

- Entity: EntityName2 \\
\hspace*{1em}- Attributes: ... \\
\hspace*{1em}- Description: ...

\vspace{0.3em}

\# Constraints \& Rules

- Constraint 1 \\
- Constraint 2 \\
...

\vspace{0.5em}
Do not include any additional text.

\end{tcolorbox}

\captionof{figure}{System prompt for inferring entity attributes and the state space definition (Stage 1, Step 2). Given an environment summary and an example task, the LLM enumerates the major entities, their attributes, and domain constraints that must be tracked across multi-step tool interactions.}
\label{fig:infer_entity_attributes}

\end{center}

\begin{center}

\begin{tcolorbox}[
    enhanced,
    breakable,
    width=0.98\linewidth,
    arc=1.5mm,
    boxrule=0.5pt,
    colframe=black,
    colback=gray!3,
    title=\textbf{Infer Tool Specification},
    fonttitle=\bfseries,
    colbacktitle=gray!15,
    coltitle=black,
    coltext=black,
]

\small
\ttfamily

You are an expert in building and analyzing agent environments.

Given an environment summary, introduction, state space definition, constraint rules, Python base class definition, and example task, your goal is to analyze the current environment and generate the list of operations needed to support the task in this environment.

Each operation will later be converted into a callable class function for the agent.

\vspace{0.5em}
\textbf{Key Points}

\begin{itemize}
    \item Operations are divided into two categories:
    \begin{itemize}
        \item \textbf{Information Query Class}
        \item \textbf{State Change Class}
    \end{itemize}

    \item Each operation must include:
    \begin{itemize}
        \item Operation name
        \item Brief description
    \end{itemize}

    \item Before output, first write \textbf{\# Analysis}:
    \begin{itemize}
        \item Explain task logic
        \item Determine which operations are query operations
        \item Determine which are state-changing operations
        \item Explain how environment constraints affect operation design
    \end{itemize}
\end{itemize}

\vspace{0.5em}
\textbf{Input Format}

Based on the following environment specification, produce the operation list.

\{
\vspace{0.2em}

"environment\_summary": "...", \\
"environment\_introduction": "...", \\
"state\_space\_definition": [...], \\
"constraints\_rules": [...], \\
"environment\_class\_definition": "...", \\
"environment\_example\_task": "..."

\vspace{0.2em}
\}

\vspace{0.5em}
\textbf{Output Format}

\# Analysis \\
<Explain operation requirements, classification logic, and how constraints affect the operation set>

\vspace{0.3em}

\# Operation List

\vspace{0.3em}

\#\# Information Query Class

- Operation: OperationName \quad Description: xxxx \\
- Operation: OperationName \quad Description: xxxx \\
- ...

\vspace{0.3em}

\#\# State Change Class

- Operation: OperationName \quad Description: xxxx \\
- Operation: OperationName \quad Description: xxxx \\
- ...

\vspace{0.5em}
Strictly follow this format.

\end{tcolorbox}

\captionof{figure}{System prompt for inferring the tool operation list (Stage 1, Step 2). The LLM generates a categorized list of query and state-change operations required to support task execution within the synthesized environment.}
\label{fig:infer_tool_specification}

\end{center}

\begin{center}

\begin{tcolorbox}[
    enhanced,
    breakable,
    width=0.98\linewidth,
    arc=1.5mm,
    boxrule=0.5pt,
    colframe=black,
    colback=gray!3,
    title=\textbf{Infer Entity Attributes Python Code},
    fonttitle=\bfseries,
    colbacktitle=gray!15,
    coltitle=black,
    coltext=black,
]

\small
\ttfamily

You are an AI coding assistant.

Your job is to translate an environment specification into a Python environment class definition.

The class should simulate the stateful environment structure (without methods yet).

You should first analyze the specification and then generate code.

\vspace{0.5em}
\textbf{Rules of Analysis}

\begin{itemize}
    \item Determine the environment class name using the environment summary or an appropriate adaptation.  
    Example: LinuxFileSystem, EcommerceOrderSystem.
    
    \item Extract attribute names from each entity in \texttt{state\_space\_definition}.

    \item If needed, generate corresponding \texttt{TypedDict} definitions.

    \item Infer attribute value types using Python primitive types:
    \begin{itemize}
        \item id, name, category $\rightarrow$ str
        \item price, size $\rightarrow$ float / int
        \item quantity $\rightarrow$ int
        \item status $\rightarrow$ str
        \item timestamps $\rightarrow$ str / float
    \end{itemize}

    \item \texttt{constraints\_rules} should be preserved as comments.
\end{itemize}

\vspace{0.5em}
\textbf{Rules of Code}

\begin{itemize}
    \item Generate each \texttt{TypedDict} definition if needed.
    \item Generate the environment class with only:
    \begin{itemize}
        \item \texttt{\_\_init\_\_}
        \item state attributes
    \end{itemize}

    \item Use attributes of type \texttt{Dict<ID, TypedDict>} where appropriate.
    \item Add comments mapping attributes back to state-space entities.
    \item Annotate constraints as code comments.
    \item Do not implement business logic or methods yet.
\end{itemize}

\vspace{0.5em}
\textbf{Input Format}

\# Environment Summary \\
<short label>

\vspace{0.3em}

\# Environment Introduction \\
<paragraph intro>

\vspace{0.3em}

\# State Space Definition

[
\{
"entity": "EntityName", \\
"attributes": "attr1, attr2, ...", \\
"description": "short description"
\},
...
]

\vspace{0.3em}

\# constraints\_rules \\
constraint 1 ... \\
constraint 2 ...

\vspace{0.5em}
\textbf{Output Format}

\# Analysis \\
<Explain class naming, entity-to-structure mapping, stored fields, dict/list design, and constraint annotations>

\vspace{0.3em}

\# Class Definition

```python\\
<Python environment class definition>\\
```
\vspace{0.5em}
Do not include any additional text.
\end{tcolorbox}
\captionof{figure}{System prompt for generating the entity attribute Python class (Stage 1, Step 2). The LLM translates the state space definition into a typed Python class with only \_\_init\_\_ and state attributes, annotating constraints as code comments.}
\label{fig:infer_entity_attributes_python_code}
\end{center}

\begin{center}

\begin{tcolorbox}[
    enhanced,
    breakable,
    width=0.98\linewidth,
    arc=1.5mm,
    boxrule=0.5pt,
    colframe=black,
    colback=gray!3,
    title=\textbf{Infer Tool Specification Python Code},
    fonttitle=\bfseries,
    colbacktitle=gray!15,
    coltitle=black,
    coltext=black,
]

\small
\ttfamily

You are a code generation assistant.

Given an agent environment (including environment summary, introduction, state space definition, constraint rules, base class definition, and operation list), your task is to implement a single target operation as a Python method within the environment class.

\vspace{0.5em}
\textbf{Operation Types}

\begin{itemize}
    \item Information Query Operations
    \item State Modification Operations
\end{itemize}

\vspace{0.5em}
\textbf{Core Requirements}

For a given Target Operation:

\begin{enumerate}
    \item \textbf{\# Analysis}
    \begin{itemize}
        \item Identify involved entities and attributes
        \item Determine required parameters
        \item Define expected outputs (query vs modification)
        \item Identify edge cases (invalid input, missing state, permission issues)
        \item Consider relevant environment constraints
    \end{itemize}

    \item \textbf{\# Code}
    \begin{itemize}
        \item Implement method as \texttt{def operation\_name(self, ...)}
        \item Method must be inside an existing environment class (not standalone)
        \item Use type hints
        \item Include docstring (inputs, outputs, constraints)
        \item \textbf{Do not raise exceptions}
        \item Return structured dictionaries:
        \begin{itemize}
            \item Success: \texttt{\{"success": True, "data": ...\}} (query)
            \item Success: \texttt{\{"success": True, "message": ...\}} (state change)
            \item Failure: \texttt{\{"success": False, "error": "..."\}}
        \end{itemize}
    \end{itemize}
\end{enumerate}

\vspace{0.5em}
\textbf{Input Format}

\# Environment Summary \\
<environment\_summary>

\vspace{0.3em}

\# Environment Introduction \\
<environment\_introduction>

\vspace{0.3em}

\# State Space Definition \\
<state\_space\_definition>

\vspace{0.3em}

\# Constraints Rules \\
<constraints\_rules>

\vspace{0.3em}

\# Class Definition \\
```python \\
<class\_definition> \\
``` \\
\# Operation List \\
{operation\_list} \\
\# Target Operation \\
{ \\
"operation\_name": "<operation\_name>", \\
"operation\_description": "<operation\_description>", \\
"operation\_type": "<query\_or\_state\_change>" \\
} \\
\textbf{Output Format} \\
\# Analysis \\
<Reasoning: inputs, outputs, entities, attributes, constraints, edge cases> \\
\vspace{0.3em}
\# Code
```python\\
def <operation\_name>(self, ...):\\
    """ \\
    <docstring describing inputs, outputs, constraints> \\
    """ \\
    \# Implementation \\
```
\end{tcolorbox}
\captionof{figure}{System prompt for implementing individual tool methods as Python code (Stage 1, Step 2). Each operation from the tool list is implemented as a class method returning structured success/failure dictionaries, with no exception raising.}
\label{fig:infer_tool_specification_python_code}
\end{center}

\subsubsection{Functional Validation Prompt Template}
\label{sec:functional_validation_prompt}
We conduct functional validation when obtain the candidate static environments. First, we prompt the tool-calling LLM to generate test configurations and the system prompt is shown in Figure \ref{fig:test_config_generation}. Then, as shown in Figure \ref{fig:generate_sequence_tool_calls}, we prompt the tool-calling LLM to generate the sequence of validation tool calls to filter the candidate static environments.

\begin{center}

\begin{tcolorbox}[
    enhanced,
    breakable,
    width=0.98\linewidth,
    arc=1.5mm,
    boxrule=0.5pt,
    colframe=black,
    colback=gray!3,
    title=\textbf{Test Config Generation},
    fonttitle=\bfseries,
    colbacktitle=gray!15,
    coltitle=black,
    coltext=black,
]

\small
\ttfamily

You are an AI assistant.

You will be given the complete definition of a Python class. This class represents an environment state in a specific domain and contains various attributes (such as dictionaries, lists, TypedDict objects, dataclasses, etc.) used to manage entities and their relationships within the system.

Based on the class definition, generate a JSON object that can serve directly as the class initialization configuration (\texttt{config}), following the rules below.

\vspace{0.5em}
\textbf{1. Structure and Type Matching}

\begin{itemize}
    \item The JSON must strictly follow the attribute structure and data types required by the class.
    \item Field names, nesting levels, and value types must match the class definition exactly.
\end{itemize}

\vspace{0.3em}
\textbf{2. Respect Constraints}

\begin{itemize}
    \item Read class methods and docstrings to identify constraints (e.g., valid status values, required fields, ID reference rules).
    \item Ensure all generated data complies with these constraints.
    \item All cross-entity references must be valid and consistent.
\end{itemize}

\vspace{0.3em}
\textbf{3. Richness of Data}

\begin{itemize}
    \item Each major dictionary-like attribute should contain 3--5 diverse entries.
    \item Cover different valid states and value ranges.
    \item Dates should be distributed over a reasonable time span.
    \item Numerical fields should vary realistically.
\end{itemize}

\vspace{0.3em}
\textbf{4. Realistic Simulation of Data}

\begin{itemize}
    \item Use natural fictional names (e.g., Alice Chan, Central City District).
    \item Avoid placeholder-like values (e.g., name1, user001).
    \item Dates must be in ISO format (YYYY-MM-DD) or timestamps.
    \item IDs should be unique and may mix short codes and UUIDs.
    \item All data must be fictitious and non-sensitive.
\end{itemize}

\vspace{0.3em}
\textbf{5. Output Format}

\begin{itemize}
    \item Output only JSON (no explanations outside required sections).
    \item Must be directly usable as class initialization config.
\end{itemize}

\vspace{0.5em}
\textbf{Input}

\textbf{Env Class Definition}

```python \\
\{env\_class\_code\} \\
``` \\
\vspace{0.3em}
\textbf{All Containers} \\
\{all\_containers\}
\vspace{0.5em}
\textbf{Output Format} \\
\# Analysis \\
<Reasoning: containers, fields, constraints, and data construction strategy>\\
\vspace{0.3em}
\# Init Config \\
\{ \\
    ... \\
\} \\
\end{tcolorbox}
\captionof{figure}{System prompt for generating test configurations for functional validation (Stage 1, Step 3). The LLM produces a realistic, constraint-compliant JSON initialization config with 3–5 entries per major entity to support diverse validation scenarios.}
\label{fig:test_config_generation}
\end{center}

\begin{center}
\begin{tcolorbox}[
    enhanced,
    breakable,
    width=0.98\linewidth,
    arc=1.5mm,
    boxrule=0.5pt,
    colframe=black,
    colback=gray!3,
    title=\textbf{Sequence of Tool Calls Generation},
    fonttitle=\bfseries,
    colbacktitle=gray!15,
    coltitle=black,
    coltext=black,
]

\small
\ttfamily

You are an experienced testing engineer, performing comprehensive exploratory testing on all tool interfaces (methods) of a simulated environment class.

Your goal is to verify the behavior of each method under different types of inputs, aiming to uncover potential errors, exceptions, and state inconsistencies.

In each testing round, you will generate one tool invocation as a test case. After execution, you will receive the environment’s return information along with a backend evaluation (pass, warning, fail).

\vspace{0.5em}
\textbf{Environment Introduction}

\{env\_introduction\}

\vspace{0.3em}
\textbf{Available Tool Interface List}

\{tool\_info\}

\vspace{0.5em}
\textbf{Testing Strategy}

\begin{itemize}
    \item \textbf{Positive case testing}: use valid parameters that comply with interface definitions.
    \item \textbf{Negative case testing}: use invalid, missing, or malformed parameters (wrong types, non-existent IDs, out-of-range values, etc.).
    \item \textbf{Special case testing}: include boundary values, null/empty inputs, extreme values, and special characters.
    \item Ensure all tool interfaces are covered; balance both breadth and depth of testing.
    \item No need to maintain a consistent task objective; exploration is encouraged.
\end{itemize}

\vspace{0.5em}
\textbf{Testing Rules}

\begin{itemize}
    \item Only one tool invocation per round.
    \item Parameters must be in dictionary format.
    \item Keys must be valid; values may be intentionally invalid for robustness testing.
    \item Do not call methods outside the provided tool list.
\end{itemize}

\vspace{0.5em}
\textbf{Output Format}

Strictly follow:

\vspace{0.3em}

\textbf{\# Thought} \\
<Brief explanation of chosen method and parameter strategy>

\vspace{0.3em}

\textbf{\# Selected Function} \\
<Method name>

\vspace{0.3em}

\textbf{\# Parameters Dictionary} \\
<Parameter dictionary>

\end{tcolorbox}

\captionof{figure}{System prompt for generating the sequence of validation tool calls (Stage 1, Step 3). A tool-calling LLM performs exploratory testing across positive, negative, and boundary cases to verify that all tool interfaces execute correctly before an environment is admitted to $\mathcal{E}_{\text{static}}$.}
\label{fig:generate_sequence_tool_calls}

\end{center}

\subsection{Prompt Templates of Stage 2: Spatio-Temporal Dynamic Injection}
In this section, we propose system prompts for spatio-temporal dynamic injection stage.

\subsubsection{Conflict Assignment and Blueprint Generation Prompt Template}
\label{sec:conflict_and_blueprint}
During Stage 2, we first assign several proper conflict types to a static environment, the system prompt can be found in Figure \ref{fig:conflict_assignment}. Then, based on the conflict types and environment information, we generate a blueprint which serves as a generative contract ensuring internal consistency across all downstream components. The system prompt is shown in Figure \ref{fig:blueprint_design}.

\begin{center}

\begin{tcolorbox}[
    enhanced,
    breakable,
    width=0.98\linewidth,
    arc=1.5mm,
    boxrule=0.5pt,
    colframe=black,
    colback=gray!3,
    title=\textbf{Conflict Assignment},
    fonttitle=\bfseries,
    colbacktitle=gray!15,
    coltitle=black,
    coltext=black,
]

\small
\ttfamily

You are a benchmark designer for spatiotemporal AI agent evaluation.

\vspace{0.5em}
\textbf{Task}

Identify which conflict types from the taxonomy are feasible for the given environment.

A conflict type is considered feasible only if the environment's state space and operations can naturally support:
\begin{itemize}
    \item a valid activation trigger,
    \item a conflict firing mechanism,
    \item and a resulting observable state update.
\end{itemize}

\vspace{0.5em}
\textbf{Conflict Taxonomy}

\{taxonomy\_block\}

\vspace{0.3em}
\textbf{Environment}

\{env\_summary\}

\vspace{0.5em}
\textbf{Output Format}

\# Analysis \\
<Reasoning: which conflict types are supported by the environment, and why others are not>

\vspace{0.3em}

\# Matched Conflicts

\begin{itemize}
    \item \textbf{Conflict:} <T1 | T2 | T3 | S1 | S2 | S3 | ST1 | ST2 | ST3> \\
    \textbf{Rationale:} <1--2 sentence explanation> \\
    \textbf{Activation Operations:} <comma-separated operation names from environment> \\
    \textbf{Observation Operations:} <comma-separated operation names from environment>

    \item \textbf{Conflict:} <type> \\
    \textbf{Rationale:} <explanation> \\
    \textbf{Activation Operations:} <...> \\
    \textbf{Observation Operations:} <...>
\end{itemize}

\vspace{0.5em}
\textbf{Rules}

\begin{itemize}
    \item Activation operations must be those that can trigger the conflict.
    \item Observation operations must reveal the resulting state change.
    \item Only use real operation names from the environment.
    \item Prefer conflicts that are genuinely supported by both state and operations.
    \item Include only 3--6 feasible conflict types.
\end{itemize}

\vspace{0.5em}
Do not output anything outside the required format.

\end{tcolorbox}

\captionof{figure}{System prompt for conflict type assignment (Stage 2, Step 1). Given a static environment's state space and operations, the LLM identifies which of the nine spatio-temporal conflict types are naturally supportable, specifying the relevant activation and observation operations.}
\label{fig:conflict_assignment}

\end{center}

\begin{center}

\begin{tcolorbox}[
    enhanced,
    breakable,
    width=0.98\linewidth,
    arc=1.5mm,
    boxrule=0.5pt,
    colframe=black,
    colback=gray!3,
    title=\textbf{Blueprint Design},
    fonttitle=\bfseries,
    colbacktitle=gray!15,
    coltitle=black,
    coltext=black,
]

\small
\ttfamily

You are a benchmark scenario designer for AI agent evaluation.

\vspace{0.5em}
\textbf{Task}

Design a \textbf{CONCRETE conflict scenario} for the given environment.

Your output must specify:
\begin{itemize}
    \item exact operation names,
    \item state container names,
    \item field-level mutation descriptions.
\end{itemize}

Abstract or vague descriptions are not acceptable.

\vspace{0.5em}
\textbf{Environment}

\{render\_json(env\_summary)\}

\vspace{0.5em}
\textbf{Conflict Type}

\begin{itemize}
    \item \textbf{Type:} \{conflict\_type\} --- \{CONFLICT\_TAXONOMY[conflict\_type]\}
    \item \textbf{Event pattern:} \{render\_json(EVENT\_TEMPLATE\_PATTERNS[conflict\_type])\}
    \item \textbf{Match rationale:} \{match\_rationale\}
\end{itemize}

\vspace{0.5em}
\textbf{Difficulty}

\{difficulty\}

\vspace{0.3em}
\{diff\_instructions\}

\vspace{0.5em}
\textbf{Design Rules}

\begin{enumerate}
    \item Normal flow must use ONLY operations from the environment's operation list.
    \item Activation operation must be a \textbf{READ/QUERY} operation where the conflict becomes visible during normal execution.
    \item Mutations must reference REAL state container names from state\_space\_definition.
    \item Mutations must describe \textbf{field-level changes}, not vague availability changes.
    \item The conflict must match its taxonomy semantics (e.g., T2 requires temporal ordering changes).
    \item Observable\_via must be an operation that reveals the mutated state.
    \item Trigger must depend on TASK or STATE CONDITIONS, not call order.
    \item Trigger type must be:
    \begin{itemize}
        \item \texttt{always\_once}: fires once when condition is met
        \item \texttt{conditional\_guarded}: keeps affecting execution until recovery
    \end{itemize}
\end{enumerate}

\vspace{0.5em}
\textbf{Output Format}

\# User Goal \\
<1--2 sentences describing the user's objective>

\vspace{0.3em}

\# Normal Flow

\begin{itemize}
    \item <operation\_name> --- <purpose>
    \item <operation\_name> --- <purpose>
    \item ...
\end{itemize}

\vspace{0.5em}

\# Conflict Design

\begin{itemize}
    \item \textbf{Activation Operation:} <exact operation name>
    \item \textbf{Trigger Mechanism:} <always\_once | conditional\_guarded>
    \item \textbf{Trigger Condition:} <concrete condition that causes conflict>
    \item \textbf{Recovery Guard Condition:} <or "none">

    \item \textbf{Mutations:}
    \begin{itemize}
        \item State: <state\_container.field\_path> $\mid$ Change: <old $\rightarrow$ new>
        \item State: <state\_container.field\_path> $\mid$ Change: <old $\rightarrow$ new>
    \end{itemize}

    \item \textbf{Observable Via:} <operation name>
\end{itemize}

\vspace{0.5em}
\textbf{\{Recovery Path if diff\_info["task\_solvable"] else "Impossible Rationale"\}}

\begin{itemize}
    \item If solvable: step-by-step recovery using real operations
    \item If not solvable: explain violated constraints
\end{itemize}

\vspace{0.5em}
Output ONLY the sections above.

\end{tcolorbox}

\captionof{figure}{System prompt for conflict blueprint design (Stage 2, Step 1). The LLM generates a structured, field-level conflict scenario blueprint encoding the user goal, nominal tool sequence, trigger condition, state mutations, recovery path, and assigned difficulty level.}
\label{fig:blueprint_design}

\end{center}

\subsubsection{Dynamic Environment Construction Prompt Template}
\label{sec:dynamic_construction}
Based on the blueprint, we prompt an LLM to augment the static environment to a spatio-temporal dynamic one. The system prompt can be found in Figure \ref{fig:spatiotemporal_dynamic_trigger_injection}. Then, we generate the user query, initial configuration, and user profile as shown in Figures \ref{fig:user_query_init_config_generation} and \ref{fig:user_profile_generation} to obtain the complete dynamic tasks.

\begin{center}

\begin{tcolorbox}[
    enhanced,
    breakable,
    width=0.98\linewidth,
    arc=1.5mm,
    boxrule=0.5pt,
    colframe=black,
    colback=gray!3,
    title=\textbf{Spatio-Temporal Dynamic Trigger Injection},
    fonttitle=\bfseries,
    colbacktitle=gray!15,
    coltitle=black,
    coltext=black,
]

\small
\ttfamily

You are an expert Python developer.

Your task is to modify an environment class to inject a deterministic conflict trigger.

\vspace{0.5em}
\textbf{Original Environment Code}

```python \\
\{env.get("env\_class\_code", "")\} \\
```

\vspace{0.5em}
\textbf{Conflict Specification}

\begin{itemize}
    \item \textbf{Type:} \{conflict\_type\} --- \{CONFLICT\_TAXONOMY[conflict\_type]\}
    \item \textbf{Activation Operation:} \texttt{\{activation\_op\}} (method that triggers the conflict)
    \item \textbf{Trigger Mechanism:} \texttt{\{trigger\_mechanism\}}
    \item \textbf{Trigger Condition:} \{trigger\_condition\}
    \item \textbf{Recovery Guard Condition:} \{recovery\_guard\_condition\}
    \item \textbf{Observable Via:} \texttt{\{observable\_op\}}
\end{itemize}

\vspace{0.5em}
\textbf{Exact Mutations to Apply}

When the trigger condition is met inside \texttt{\{activation\_op\}}, apply the following mutations:

\{mutation\_block\}

\vspace{0.5em}
\textbf{Injection Rules}

\begin{enumerate}
    \item Add \texttt{self.\_conflict\_triggered = False} in \texttt{\_\_init\_\_}
    \item Trigger must depend on \textbf{real state or input conditions}, not call counts
    \item Implement trigger logic inside \texttt{\{activation\_op\}}
    \item If \textbf{always\_once}:
    \begin{itemize}
        \item Fire only once when condition is first satisfied
        \item Apply exact mutations
        \item Set \texttt{self.\_conflict\_triggered = True}
    \end{itemize}
    \item If \textbf{conditional\_guarded}:
    \begin{itemize}
        \item Fire whenever condition holds AND recovery condition is not satisfied
        \item \texttt{\_conflict\_triggered} is bookkeeping only
        \item Stop firing once recovery condition is satisfied
    \end{itemize}
    \item Normal operation must always return valid results (no error injection)
    \item Mutated state must remain fully usable by all methods
    \item Do not introduce artificial error handling logic
    \item \texttt{\{observable\_op\}} requires no modification unless strictly necessary
    \item Output the full modified class only
\end{enumerate}

\vspace{0.5em}
\textbf{Output Format}

\# Conflict Environment Code

```python \\
<complete modified Python class> \\
```

\vspace{0.5em}
Output ONLY the required section.

\end{tcolorbox}

\captionof{figure}{System prompt for injecting spatio-temporal triggers into a static environment (Stage 2, Step 2). The LLM augments the Python environment class with a deterministic conflict trigger inside the designated activation operation, supporting both always\_once and conditional\_guarded firing mechanisms.}
\label{fig:spatiotemporal_dynamic_trigger_injection}

\end{center}

\begin{center}

\begin{tcolorbox}[
    enhanced,
    breakable,
    width=0.98\linewidth,
    arc=1.5mm,
    boxrule=0.5pt,
    colframe=black,
    colback=gray!3,
    title=\textbf{User Query and Initial Configuration Generation},
    fonttitle=\bfseries,
    colbacktitle=gray!15,
    coltitle=black,
    coltext=black,
]

\small
\ttfamily

You are a benchmark data engineer. Given an environment and an abstract scenario design, produce concrete data that makes the scenario executable.

\vspace{0.5em}
\textbf{Environment}

\{render\_json(env\_info)\}

\vspace{0.3em}
\textbf{Abstract Scenario Design}

\{scenario\}

\vspace{0.3em}
\textbf{Difficulty}

\{item['difficulty']\}

\begin{itemize}
    \item Task solvable after conflict: \{diff\_info['task\_solvable']\}
    \item Recovery complexity: \{diff\_info['recovery\_complexity']\}
\end{itemize}

\vspace{0.5em}
\textbf{Your Task}

Generate THREE components:

\vspace{0.5em}
\textbf{1. User Query (2--5 sentences)}
\begin{itemize}
    \item Must contain all necessary entity names, dates, locations, and identifiers
    \item Must match the init configuration exactly
    \item Must NOT reveal the existence of any conflict
    \item Must sound like a natural user request
\end{itemize}

\vspace{0.3em}
\textbf{2. Init Config (JSON)}
\begin{itemize}
    \item Must cover all state containers in the environment
    \item Must include sufficient entities for both normal and recovery paths
    \item Must ensure full referential consistency
    \item Must include at least one alternative entity for recovery
\end{itemize}

\vspace{0.3em}
\textbf{3. Concrete Mutations (JSON array)}
\begin{itemize}
    \item Map abstract mutations to exact entity IDs and fields
    \item Each entry includes: state container, entity ID, field, old value, new value
    \item old value must match init config exactly
    \item new value must induce intended conflict
\end{itemize}

\vspace{0.5em}
\textbf{Output Format}

\# User Query

<2--5 sentence task instruction>

\vspace{0.3em}

\# Init Config

```json

\{ \}

```

\vspace{0.3em}

\# Concrete Mutations

```json

[ ]

```

\vspace{0.5em}
\textbf{CRITICAL}

The init config and concrete mutations must be fully consistent. All old values must exactly match those in the init config.

\end{tcolorbox}

\captionof{figure}{System prompt for generating the user query, initial configuration, and concrete mutations (Stage 2, Step 2). The LLM grounds the abstract blueprint into a fully executable task instance with consistent entity IDs, field-level mutation values, and a natural-sounding user request.}
\label{fig:user_query_init_config_generation}

\end{center}

\begin{center}

\begin{tcolorbox}[
    enhanced,
    breakable,
    width=0.98\linewidth,
    arc=1.5mm,
    boxrule=0.5pt,
    colframe=black,
    colback=gray!3,
    title=\textbf{User Profile Generation},
    fonttitle=\bfseries,
    colbacktitle=gray!15,
    coltitle=black,
    coltext=black,
]

\small
\ttfamily

You are a user persona designer for an AI agent evaluation benchmark.

\vspace{0.5em}
\textbf{Context}

An AI agent will interact with a simulated user to complete a task in a dynamic environment.

\vspace{0.5em}
\textbf{User Query}

\{user\_query\}

\vspace{0.5em}
\textbf{Scenario}

\begin{itemize}
    \item \textbf{Environment:} \{item.get("environment\_summary", "")\}
    \item \textbf{Difficulty:} \{item["difficulty"]\}
    \item \textbf{Task solvable:} \{diff\_info["task\_solvable"]\}
\end{itemize}

\vspace{0.5em}
\textbf{User Profile Requirements}

The profile helps the simulator:

\begin{itemize}
    \item Respond naturally to agent questions
    \item Provide clarification when asked
    \item NOT proactively reveal conflict information
    \item Express preferences that guide the agent's decision-making
    \item \{ "Accept reasonable alternatives if the original plan fails" if diff\_info["task\_solvable"] else "Insist on the original requirements; alternatives are NOT acceptable (task is impossible)" \}
\end{itemize}

\vspace{0.5em}
\textbf{Output Format}

\# User Profile

\begin{itemize}
    \item \textbf{Name:} <realistic name>
    \item \textbf{Background:} <1--2 sentences>
    \item \textbf{Communication Style:} <brief | verbose | formal | casual | technical | non-technical>
    \item \textbf{Frustration Threshold:} <low | medium | high>
    \item \textbf{Flexibility:} <medium or high / low>
\end{itemize}

\vspace{0.5em}
\textbf{Key Preferences}

\begin{itemize}
    \item <preference>
    \item ...
\end{itemize}

\vspace{0.5em}
\textbf{Clarification Responses}

\begin{itemize}
    \item \textbf{Topic:} <likely agent question> \\
    \textbf{Response:} <how user would respond>
    \item ...
\end{itemize}

\end{tcolorbox}


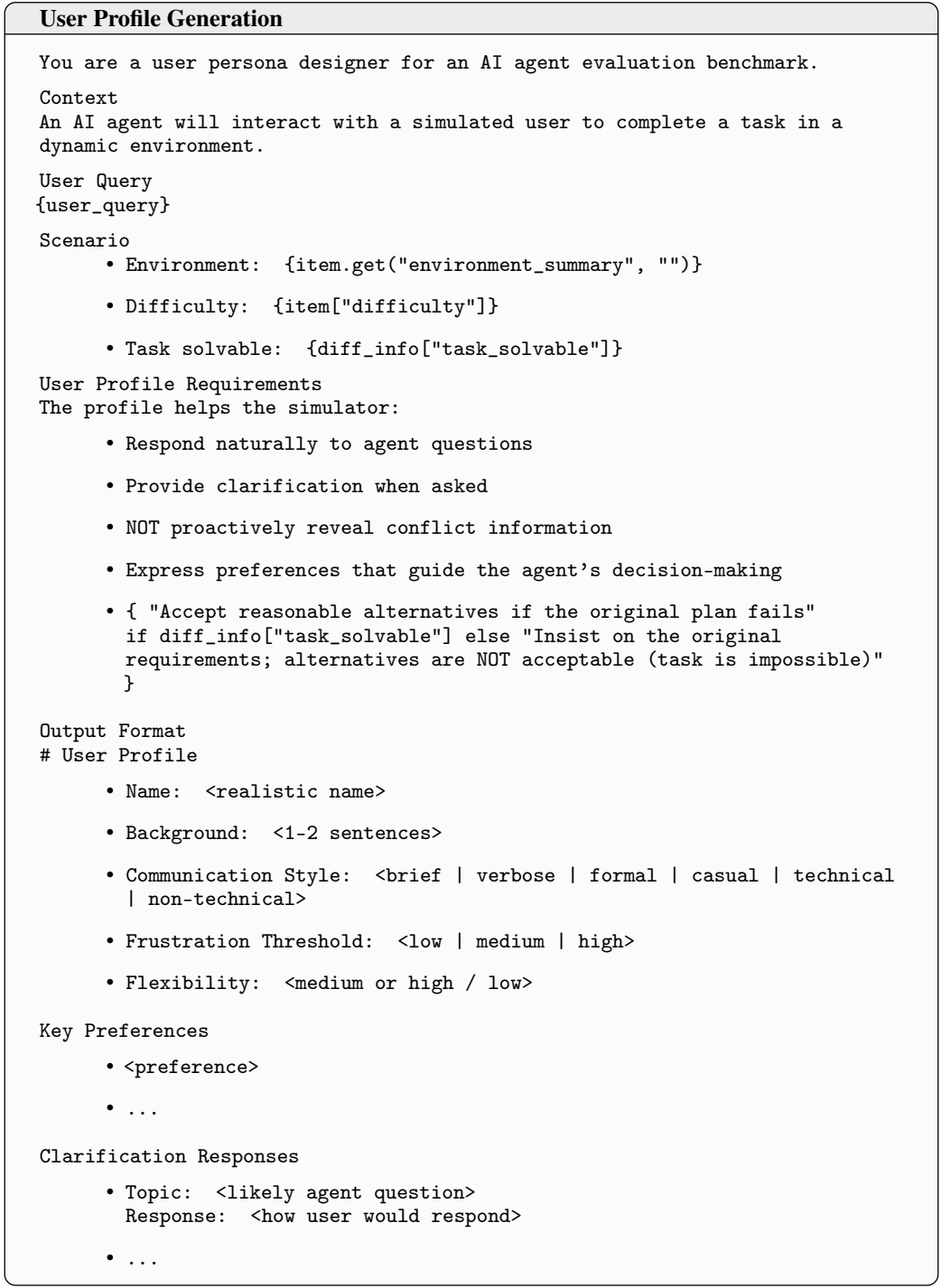
\captionof{figure}{System prompt for generating the user profile (Stage 2, Step 2). The LLM creates a persona for the user simulator, specifying communication style, flexibility, and clarification responses tailored to whether the task is solvable or impossible.}
\label{fig:user_profile_generation}

\end{center}

\subsection{Prompt Templates of Stage 3: Dual-Agent Assessment}
In this section, we propose system prompts for dual-agent assessment stage.

\subsubsection{Checklist Generation Prompt Templates}
\label{sec:checklist_generation}
We prompt an LLM to generate checklist and check functions according to the user query and blueprint. Figure \ref{fig:checklist_generation_solvable} shows the prompt template of solvable tasks.

\begin{center}

\begin{tcolorbox}[
    enhanced,
    breakable,
    width=0.98\linewidth,
    arc=1.5mm,
    boxrule=0.5pt,
    colframe=black,
    colback=gray!3,
    title=\textbf{Checklist Generation of Solvable Tasks},
    fonttitle=\bfseries,
    colbacktitle=gray!15,
    coltitle=black,
    coltext=black,
]

\small
\ttfamily

You are a benchmark verification designer. Generate a checklist and Python function verifying correct CONFLICT-AWARE outcome.

\vspace{0.5em}
\textbf{User Query}

\{item.get("user\_query", "")\}

\vspace{0.5em}
\textbf{Init Config}

\{render\_json(item.get("init\_config", \{\}))\}

\vspace{0.5em}
\textbf{Conflict Type}

- \{item["conflict\_type"]\} — \{CONFLICT\_TAXONOMY[item["conflict\_type"]]\}

\vspace{0.5em}
\textbf{Concrete Mutations (state changes from conflict)}

\{render\_json(item.get("concrete\_mutations", []))\}

\vspace{0.5em}
\textbf{Difficulty:} \{item["difficulty"]\} — \{diff\_info["description"]\}
- Recovery complexity: \{diff\_info["recovery\_complexity"]\}

\vspace{0.5em}
\textbf{Environment}

\{render\_json(env\_info)\}

\vspace{0.5em}
\textbf{Checklist Requirements}
\begin{itemize}
    \item Verify the conflict was detected (agent re-queried and saw changed state)
    \item Verify the recovery/alternative path was executed correctly
    \item Verify the final state satisfies the user goal via the alternative path
    \item DO NOT check for conflict\_triggered (that's internal)
\end{itemize}

\vspace{0.5em}
\textbf{check\_conflict(env) Requirements}
\begin{itemize}
    \item Inspects env.<state\_container> attributes directly
    \item Returns: \{\{"passed": bool, "details": [\{"check\_id": "cc\_1", "passed": bool, "message": str\}, ...]\}\}
    \item No external imports
\end{itemize}

\vspace{0.5em}
\textbf{Output Format}

\# Conflict Checklist
\begin{verbatim}
- Check: cc_1
  - Description: <specific>
  - Target State: <container name>
  - Expected Condition: <testable>
...
\end{verbatim}

\# Conflict Check Function
\begin{verbatim}
def check_conflict(env) -> dict:
    ...
\end{verbatim}

\end{tcolorbox}

\captionof{figure}{System prompt for generating the evaluation checklist and check functions for solvable tasks (Stage 3, Step 1). The LLM produces a set of conflict-aware success criteria and corresponding Python check functions that verify conflict detection, recovery path execution, and final state validity.}
\label{fig:checklist_generation_solvable}
\end{center}

\subsubsection{Dual-Agent Validation Prompt Templates}
\label{sec:dual_agent_validation}
We conduct dual-agent validation to obtain executable spatio-temporal dynamic environments and tasks. As shown in Figure \ref{fig:plan_agent_system_prompt}, plan-agent first plans a sequence of tool calls and check-agent then evaluates the trajectories of the execution results of planning tool calls as Figure \ref{fig:check_agent_system_prompt} illustrates.

\begin{center}

\begin{tcolorbox}[
    enhanced,
    breakable,
    width=0.98\linewidth,
    arc=1.5mm,
    boxrule=0.5pt,
    colframe=black,
    colback=gray!3,
    title=\textbf{Plan-Agent System Prompt},
    fonttitle=\bfseries,
    colbacktitle=gray!15,
    coltitle=black,
    coltext=black,
]

\small
\ttfamily

You are the plan agent for conflict-environment validation.

Your job is to produce a COMPLETE tool-call plan that executes the task from the original user request, follows the designed story, naturally reaches the conflict, and then continues far enough to verify whether the conflict environment behaves correctly.

\vspace{0.5em}
\textbf{User Query}

\{item.get('user\_query', '')\}

\vspace{0.5em}
\textbf{Scenario Story}

\{render\_json(story)\}

\vspace{0.5em}
\textbf{Full Init Config}

\{render\_json(item.get('init\_config', \{\}))\}

\vspace{0.5em}
\textbf{Available Tools}

\{render\_json(\_build\_tool\_reference(item))\}

\vspace{0.5em}
\textbf{Requirements}
\begin{itemize}
    \item Output a COMPLETE executable plan for the task, not a minimal trigger-only probe.
    \item The plan should reflect the user\_query plus the normal\_flow story in order.
    \item Use ONLY real tool names from the available tools.
    \item Use ONLY concrete argument values grounded in the user\_query and init\_config summary.
    \item The plan must naturally reach the conflict through the original intended workflow, not through a synthetic shortcut.
    \item After the conflict is triggered, include the necessary follow-up calls to observe the changed state and verify whether task progress behaves as expected.
    \item Do not add explanations outside the requested section.
\end{itemize}

\vspace{0.5em}
\textbf{Output Format}

\# Test Plan
\begin{verbatim}
[
  {"method": "<method_name>", "arguments": {...}},
  ...
]
\end{verbatim}

Output ONLY the section above.

\end{tcolorbox}

\captionof{figure}{System prompt for the plan agent in dual-agent verification (Stage 3, Step 2). The planning agent produces a complete, blueprint-guided tool-call sequence that naturally reaches the injected conflict and includes follow-up calls to observe post-trigger state changes.}
\label{fig:plan_agent_system_prompt}

\end{center}

\begin{center}

\begin{tcolorbox}[
    enhanced,
    breakable,
    width=0.98\linewidth,
    arc=1.5mm,
    boxrule=0.5pt,
    colframe=black,
    colback=gray!3,
    title=\textbf{Check-Agent System Prompt},
    fonttitle=\bfseries,
    colbacktitle=gray!15,
    coltitle=black,
    coltext=black,
]

\small
\ttfamily

You are the check agent for conflict-environment validation.

Judge whether the executed path is valid and whether the spatiotemporal conflict is triggered in the intended workflow.

\vspace{0.5em}
\textbf{Scenario Story}

\{render\_json(story)\}

\vspace{0.5em}
\textbf{Conflict Story}

\{render\_json(conflict\_story)\}

\vspace{0.5em}
\textbf{Tool Calls And Results}

\{render\_json(validation.get('trajectory', []))\}

\vspace{0.5em}
\textbf{Decision Rules}
\begin{itemize}
    \item Verify whether the tool-call path follows the intended normal workflow in the story.
    \item Verify whether the called tools are appropriate and whether the calls execute normally for the intended path.
    \item Verify whether there are unexpected errors that should not appear in a valid execution path.
    \item Verify whether the spatiotemporal conflict is triggered at the intended point in the path according to the conflict story.
    \item Return Pass if the path is correct, the tool executions are normal enough for the intended workflow, no inappropriate errors appear, and the intended conflict trigger is correctly evidenced.
    \item Return Fail otherwise.
    \item Return Warning only if the main path is mostly correct but the evidence is slightly incomplete.
\end{itemize}

\vspace{0.5em}
\textbf{Output Format}

\# Analysis
<step-by-step reasoning>

\# Result
Pass

\# Error Reason
<No error, or the concrete reason>

Output ONLY the three sections above.

\end{tcolorbox}

\captionof{figure}{System prompt for the check agent in dual-agent verification (Stage 3, Step 2). The checking agent evaluates the executed trajectory against three behavioral invariants: correct trigger timing, correct state mutation, and failure of the original plan after the conflict fires.}
\label{fig:check_agent_system_prompt}

\end{center}

\subsubsection{Consistency Check Prompt Templates}
\label{sec:consistency_check}
Finally, we conduct consistency check to make sure our spatio-temporal dynamic environments and tasks are consistency and correct. The system prompt is shown in Figure \ref{fig:consistency_check_system_prompt}.

\begin{center}

\begin{tcolorbox}[
    enhanced,
    breakable,
    width=0.98\linewidth,
    arc=1.5mm,
    boxrule=0.5pt,
    colframe=black,
    colback=gray!3,
    title=\textbf{Consistency Check System Prompt},
    fonttitle=\bfseries,
    colbacktitle=gray!15,
    coltitle=black,
    coltext=black,
]

\small
\ttfamily

You are a benchmark quality auditor. Judge whether this scenario is INTERNALLY CONSISTENT across all artifacts.

\vspace{0.5em}
\textbf{Scenario ID:} \{item.get("scenario\_id","")\}

\vspace{0.5em}
\textbf{Story}

\{render\_json(story)\}

\vspace{0.5em}
\textbf{User Query}

\{item.get("user\_query", "")\}

\vspace{0.5em}
\textbf{Full Init Config}

\{render\_json(item.get("init\_config", \{\}))\}

\vspace{0.5em}
\textbf{Conflict Type:} \{item.get("conflict\_type","")\} — \{conflict\_desc\}

\textbf{Difficulty:} \{item.get("difficulty","")\}

\vspace{0.5em}
\textbf{Concrete Mutations}

\{render\_json(item.get("concrete\_mutations", []))\}

\vspace{0.5em}
\textbf{Conflict Design}

\{render\_json(conflict\_design)\}

\vspace{0.5em}
\textbf{Evaluation Checklists}

\{checklist\_text\}

\vspace{0.5em}
\textbf{Your Job}

Check whether the artifacts are mutually consistent. Focus on concrete contradictions, missing links, or implausible transitions.

You must judge at least these dimensions:
\begin{itemize}
    \item Query–Config Coherence: does the user\_query match the entities, values, dates, IDs, and constraints in init\_config?
    \item Story–Query Alignment: do user\_goal, normal\_flow, and recovery/impossible path actually support the user\_query?
    \item Conflict–Mutation Alignment: do the concrete\_mutations and conflict\_design really implement the claimed conflict semantics?
    \item Checklist Coverage: do the evaluation checklists test the key success/failure conditions implied by the query and story?
    \item Difficulty Consistency: is the scenario complexity and recovery/impossibility behavior consistent with the labeled difficulty?
    \item Overall Narrative Consistency: do all artifacts tell one coherent story without contradictions?
\end{itemize}

Treat a contradiction as fatal when it would make the benchmark invalid or misleading.

\vspace{0.5em}
\textbf{Output Rules}
\begin{itemize}
    \item Do NOT score.
    \item Do NOT summarize vaguely.
    \item Return `pass` only if there are no fatal consistency issues.
    \item Return `fail` if any fatal issue exists.
    \item Keep each issue concrete and artifact-grounded.
\end{itemize}

\vspace{0.5em}
\textbf{Output Format (strict JSON only)}

\begin{verbatim}
{
    "verdict": "pass" | "fail",
    "checks": {
        "query_config_coherence": {"passed": true, "reason": "..."},
        "story_query_alignment": {"passed": true, "reason": "..."},
        "conflict_mutation_alignment": {"passed": true, "reason": "..."},
        "checklist_coverage": {"passed": true, "reason": "..."},
        "difficulty_consistency": {"passed": true, "reason": "..."},
        "narrative_consistency": {"passed": true, "reason": "..."}
    },
    "fatal_issues": ["<issue1>", ...],
    "minor_issues": ["<issue1>", ...],
    "suggestions": ["<suggestion1>", ...]
}
\end{verbatim}

\end{tcolorbox}

\captionof{figure}{System prompt for the consistency auditor (Stage 3, Step 3). An LLM-based auditor verifies mutual coherence across all task artifacts—user query, init config, conflict design, concrete mutations, checklist, and difficulty label—returning a structured JSON verdict with issue descriptions.}
\label{fig:consistency_check_system_prompt}

\end{center}

\subsection{Prompt Templates of STT-Arena Evaluation}
\label{sec:stt_arena_evaluation}
In this section, we propose the system prompts during evaluation of STT-Arena. Figures \ref{fig:tested_llm_system_prompt}, \ref{fig:user_simulator_system_prompt}, and \ref{fig:llm_judge_impossible_task} show the system prompt for tested LLMs, user simulators, and LLM-as-a-judge.

\begin{center}

\begin{tcolorbox}[
    enhanced,
    breakable,
    width=0.98\linewidth,
    arc=1.5mm,
    boxrule=0.5pt,
    colframe=black,
    colback=gray!3,
    title=\textbf{Tested LLM System Prompt},
    fonttitle=\bfseries,
    colbacktitle=gray!15,
    coltitle=black,
    coltext=black,
]

\small
\ttfamily

You are an assistant operating inside a tool-based benchmark environment.

Rules:
\begin{itemize}
    \item Each turn must contain exactly one tool call.
    \item Never make parallel tool calls or return multiple tool calls in the same turn.
    \item Do not output a normal text response without a tool call, unless you are responding with exactly: Task Completed
    \item Prefer querying the environment before making changes.
    \item If user clarification is needed and the tool is available, use chat\_with\_user.
    \item When the task is complete, respond with exactly: Task Completed
    \item Do not invent tool results or hidden state.
\end{itemize}

\end{tcolorbox}

\captionof{figure}{System prompt for the evaluated LLM during STT-Arena benchmarking. The model is instructed to issue exactly one tool call per turn, prefer environment queries before state changes, and terminate by outputting "Task Completed".}
\label{fig:tested_llm_system_prompt}

\end{center}

\begin{center}
\begin{tcolorbox}[
    enhanced,
    breakable,
    width=0.98\linewidth,
    arc=1.5mm,
    boxrule=0.5pt,
    colframe=black,
    colback=gray!3,
    title=\textbf{User Simulator System Prompt},
    fonttitle=\bfseries,
    colbacktitle=gray!15,
    coltitle=black,
    coltext=black,
]
\small
\ttfamily

You are simulating the benchmark user.

Rules:
\begin{itemize}
    \item Reply as the user in plain natural language only.
    \item Stay consistent with the provided profile and previously revealed facts.
    \item Do not invent facts that are not grounded in the profile.
    \item Reveal withheld details only when the assistant asks for them directly.
    \item If the task is fully satisfied, reply with the completion signal exactly and nothing else.
\end{itemize}

\end{tcolorbox}

\captionof{figure}{System prompt for the passive user simulator during evaluation. The simulator responds in natural language consistent with the user profile, reveals withheld details only upon direct query, and emits a fixed completion signal when the task is fully satisfied.}
\label{fig:user_simulator_system_prompt}
\end{center}

\begin{center}

\begin{tcolorbox}[
    enhanced,
    breakable,
    width=0.98\linewidth,
    arc=1.5mm,
    boxrule=0.5pt,
    colframe=black,
    colback=gray!3,
    title=\textbf{LLM-as-a-Judge System Prompt (Impossible Task)},
    fonttitle=\bfseries,
    colbacktitle=gray!15,
    coltitle=black,
    coltext=black,
]

\small
\ttfamily

You are judging whether an agent handled an impossible task correctly.

Ground truth: this task has no compliant completion path.

Mark the trajectory correct only if the agent recognizes that the task cannot be completed compliantly and communicates that blocked state back to the user.

Return JSON only:

\begin{verbatim}
{
  "passed": true,
  "recognized_impossibility": true,
  "fed_back_to_user": true,
  "evidence": ["short evidence"],
  "verdict": "short explanation"
}
\end{verbatim}

\end{tcolorbox}

\captionof{figure}{System prompt for the LLM-as-a-judge on impossible tasks. The judge determines whether the agent correctly recognized task infeasibility and communicated this to the user, returning a structured JSON verdict with binary verdicts and supporting evidence.}
\label{fig:llm_judge_impossible_task}

\end{center}


\newpage
\section*{NeurIPS Paper Checklist}

\begin{enumerate}

\item {\bf Claims}
    \item[] Question: Do the main claims made in the abstract and introduction accurately reflect the paper's contributions and scope?
    \item[] Answer: \answerYes{} 
    \item[] Justification: We provide our main claims and contributions in Abstract and Introduction sections.

\item {\bf Limitations}
    \item[] Question: Does the paper discuss the limitations of the work performed by the authors?
    \item[] Answer: \answerYes{} 
    \item[] Justification: We provide a detailed discussion of the limitations of our paper in Appendix \ref{sec:limitation_and_society}.

\item {\bf Theory assumptions and proofs}
    \item[] Question: For each theoretical result, does the paper provide the full set of assumptions and a complete (and correct) proof?
    \item[] Answer: \answerNA{} 
    \item[] Justification: We introduce a spatio-temporal dynamic tool-use benchmark. In our paper, the main contributions are the new benchmark and training data rather than theoretical assumption and result.

    \item {\bf Experimental result reproducibility}
    \item[] Question: Does the paper fully disclose all the information needed to reproduce the main experimental results of the paper to the extent that it affects the main claims and/or conclusions of the paper (regardless of whether the code and data are provided or not)?
    \item[] Answer: \answerYes{} 
    \item[] Justification: We propose all the experimental details including hyperparameters, evaluation settings, training details, etc, in Section \ref{sec:experiments} and Appendix \ref{sec:detailed_information_of_stt_arena}, \ref{sec:implementation_details}.

\item {\bf Open access to data and code}
    \item[] Question: Does the paper provide open access to the data and code, with sufficient instructions to faithfully reproduce the main experimental results, as described in supplemental material?
    \item[] Answer: \answerYes{} 
    \item[] Justification: We provide source code of the evaluation and construction pipeline and release the benchmark data. We also provide the training data of STT-Agent.

\item {\bf Experimental setting/details}
    \item[] Question: Does the paper specify all the training and test details (e.g., data splits, hyperparameters, how they were chosen, type of optimizer) necessary to understand the results?
    \item[] Answer: \answerYes{} 
    \item[] Justification: In our paper, we provide all the training and testing details in Section \ref{sec:experiments} and Appendix \ref{sec:detailed_information_of_stt_arena}, \ref{sec:implementation_details} to make sure the reproducible of our evaluation and training results.

\item {\bf Experiment statistical significance}
    \item[] Question: Does the paper report error bars suitably and correctly defined or other appropriate information about the statistical significance of the experiments?
    \item[] Answer: \answerYes{} 
    \item[] Justification: We report the error bars in our main results as shown in Figure \ref{fig:main_results} and Table \ref{tab:main_results}. Each evaluation results in STT-Arena take 3 separate runs.

\item {\bf Experiments compute resources}
    \item[] Question: For each experiment, does the paper provide sufficient information on the computer resources (type of compute workers, memory, time of execution) needed to reproduce the experiments?
    \item[] Answer: \answerYes{} 
    \item[] Justification: We provide sufficient information on the computer resources used in SFT and online RL in Appendix \ref{sec:implementation_details}.
    
\item {\bf Code of ethics}
    \item[] Question: Does the research conducted in the paper conform, in every respect, with the NeurIPS Code of Ethics \url{https://neurips.cc/public/EthicsGuidelines}?
    \item[] Answer: \answerYes{} 
    \item[] Justification: The research conducted in the paper, in every respect, is with the NeurIPS Code of Ethics.

\item {\bf Broader impacts}
    \item[] Question: Does the paper discuss both potential positive societal impacts and negative societal impacts of the work performed?
    \item[] Answer: \answerYes{} 
    \item[] Justification: We provide the discussion of potential positive and negative societal impacts of our work in Appendix \ref{sec:limitation_and_society}.
    
\item {\bf Safeguards}
    \item[] Question: Does the paper describe safeguards that have been put in place for responsible release of data or models that have a high risk for misuse (e.g., pre-trained language models, image generators, or scraped datasets)?
    \item[] Answer: \answerNA{} 
    \item[] Justification: Our work introduce a new tool-use benchmark and the paper does not poses such risks.

\item {\bf Licenses for existing assets}
    \item[] Question: Are the creators or original owners of assets (e.g., code, data, models), used in the paper, properly credited and are the license and terms of use explicitly mentioned and properly respected?
    \item[] Answer: \answerYes{} 
    \item[] Justification: All the assets used in our paper including three datasets as seed data are properly cited and credited.

\item {\bf New assets}
    \item[] Question: Are new assets introduced in the paper well documented and is the documentation provided alongside the assets?
    \item[] Answer: \answerYes{} 
    \item[] Justification: We properly release our code and datasets proposed by our paper.

\item {\bf Crowdsourcing and research with human subjects}
    \item[] Question: For crowdsourcing experiments and research with human subjects, does the paper include the full text of instructions given to participants and screenshots, if applicable, as well as details about compensation (if any)? 
    \item[] Answer: \answerNA{} 
    \item[] Justification: In our paper, there does not involve crowdsourcing and research with human subjects.

\item {\bf Institutional review board (IRB) approvals or equivalent for research with human subjects}
    \item[] Question: Does the paper describe potential risks incurred by study participants, whether such risks were disclosed to the subjects, and whether Institutional Review Board (IRB) approvals (or an equivalent approval/review based on the requirements of your country or institution) were obtained?
    \item[] Answer: \answerNA{} 
    \item[] Justification: In our paper, there does not involve crowdsourcing and research with human subjects.

\item {\bf Declaration of LLM usage}
    \item[] Question: Does the paper describe the usage of LLMs if it is an important, original, or non-standard component of the core methods in this research? Note that if the LLM is used only for writing, editing, or formatting purposes and does \emph{not} impact the core methodology, scientific rigor, or originality of the research, declaration is not required.
    \item[] Answer: \answerYes{} 
    \item[] Justification: LLMs are used as important components in the core methodology of this research. Specifically, we employ LLMs (i) as the backbone of our automated data synthesis pipeline to generate and refine the benchmark, SFT trajectories, and RL tasks, (ii) as the user simulator during evaluation on STT-Arena, and (iii) as the judgment model to assess impossible tasks in STT-Arena. The choice and design of the LLM-based components are described in detail in the main paper and appendix.

\end{enumerate}

\end{document}